\documentclass{article} %
\usepackage[preprint]{neurips_2026}   %

\PassOptionsToPackage{hyphens}{url} %
\usepackage{hyperref}
\usepackage{url}

\usepackage[T1]{fontenc}
\usepackage{booktabs}
\usepackage{amsfonts}
\usepackage{nicefrac}
\usepackage{microtype}
\usepackage{xcolor} %

\usepackage{float}
\usepackage{graphicx}
\usepackage{caption}
\usepackage{subcaption}

\usepackage{amsmath}
\usepackage{amssymb}
\usepackage{mathtools}

\usepackage{amsthm}
\theoremstyle{plain}
\newtheorem{theorem}{Theorem}[section]
\newtheorem{proposition}[theorem]{Proposition}
\newtheorem{lemma}[theorem]{Lemma}
\newtheorem{corollary}[theorem]{Corollary}
\theoremstyle{definition}

\theoremstyle{remark}

\theoremstyle{plain}
\newtheorem*{propositionrep}{Proposition~\ref{prop:realized-state}}
\newtheorem*{theoremrep}{Theorem~\ref{thm:nbf-consistency}}
\newtheorem*{corollaryrep}{Corollary~\ref{cor:nbf-rate}}

\usepackage[ruled,vlined,linesnumbered]{algorithm2e}

\usepackage{tikz}
\usetikzlibrary{arrows.meta,positioning,calc,backgrounds,fit,shapes.geometric}

\renewcommand{\cite}[1]{{\color{red}USE CITEP OR CITET}}

\newcommand{\mc}{\mathcal}
\newcommand{\mb}{\mathbb}
\newcommand{\reals}{\mb R}
\newcommand{\R}{\reals}
\newcommand{\E}{\mathbb{E}}

\newcommand{\defeq}{\ensuremath{\stackrel{\rm def}{=}}}
\newcommand{\inftynorm}[1]{\left\lVert#1\right\rVert_\infty}     %

\newcommand{\iidsample}{\stackrel{\text{iid}}{\sim}}
\newcommand{\asconv}{\stackrel{\text{a.s.}}{\longrightarrow}}

\newcommand{\state}{x}
\newcommand{\statespace}{X}

\newcommand{\transitionfunc}{T}

\newcommand{\obs}{y}
\newcommand{\obsspace}{Y}
\newcommand{\obsfunc}{\mc Y}

\newcommand{\jointpolicy}{\pi}

\newcommand{\action}{a}
\newcommand{\actionspace}{A}

\newcommand{\env}{G}

\newcommand{\initialstatedist}{b_0}

\newcommand{\kernel}{K_{\obs}}

\newcommand{\history}{h}
\newcommand{\statetime}[1]{\state^{(#1)}}

\newcommand{\obstime}[1]{\obs^{(#1)}}

\newcommand{\bdist}{b(\state)}
\newcommand{\bdisttime}{b_t(\state)}
\newcommand{\bdistparams}{\theta}
\newcommand{\bdistapprox}{b_\bdistparams(\state)}
\newcommand{\bsamples}{\state_{1:n}}
\newcommand{\bsamplesexpanded}{(\state_1, \state_2, \dots, \state_n)}
\newcommand{\bembed}{\bdistparams}

\newcommand{\embedfuncparams}{\phi}

\newcommand{\embedfunc}{\mc E_\embedfuncparams}

\newcommand{\generator}{\mc G_\flowparams}

\newcommand{\flowparams}{\psi}

\newcommand{\prior}{p(z)}

\newcommand{\shortloss}{\mc L(\embedfuncparams, \flowparams)}

\newcommand{\particle}{\state_i}
\newcommand{\particleset}{\bsamples}
\newcommand{\particleweight}{w_i}
\newcommand{\weightvec}{w_{1:n}}

\newcommand{\T}{t_{\text{max}}}

\newcommand{\testfunction}{\varphi}

\newcommand{\ttargetexpectation}{\E_{b_t}[\testfunction]}
\newcommand{\testimate}{\hat{\mu}_t^{(n)}(\testfunction)}
\newcommand{\estimate}{\hat{\mu}^{(n)}(\testfunction)}
\newcommand{\tnextestimate}{\hat{\mu}_{t+1}^{(n)}
(\testfunction)}

\newcommand{\snisfullprime}{\frac{\sum_{i=1}^n \particleweight \testfunction(\particle^\prime)}{\sum_{i=1}^n \particleweight}}
\newcommand{\absdiff}{|\testimate - \ttargetexpectation|}
\newcommand{\maxabsdiff}{\sup_{0 \leq t \leq \T} \absdiff}
\newcommand{\proposal}{q}
\newcommand{\weightfunction}{W}

\newcommand{\denominator}{B_n}

\newcommand{\erri}{Z_i}

\NewDocumentCommand{\markovstate}{o}{
  \history\IfValueT{#1}{_{#1}}
}

\newcommand{\weightedparticleset}{S}

\title{Neural Bayesian Filtering}

\author{%
Christopher Solinas\textsuperscript{1, $\dagger$},
Radovan Halu\v{s}ka\textsuperscript{3},
David Sychrovsk\'{y}\textsuperscript{3},
Finbarr Timbers,\\
\textbf{Nolan Bard\textsuperscript{5},
Michael Buro\textsuperscript{1},
Martin Schmid\textsuperscript{3,4},
Nathan R. Sturtevant\textsuperscript{1,2},
Michael Bowling\textsuperscript{1,2}} \\
\textsuperscript{1}University of Alberta,
\textsuperscript{2}Alberta Machine Intelligence Institute (Amii) \\
\textsuperscript{3}Charles University, Prague,
\textsuperscript{4}EquiLibre Technologies, Inc.,
\textsuperscript{5}Sony AI \\
\texttt{\{\textsuperscript{$\dagger$}solinas,mburo,nathanst,mbowling\}@ualberta.ca}, \\
\texttt{\{haluskr,sychrovsky,schmidm\}@kam.mff.cuni.cz},
\texttt{nolan.bard@sony.com}
}

\begin{document}

\maketitle

\begin{abstract}
Sequential estimation under partial observability requires tracking beliefs that may be high-dimensional, multimodal, and non-Gaussian.
Classical Bayesian filters generalize zero-shot to any system whose dynamics can be evaluated, but their representations scale poorly: parametric filters struggle to capture multimodality, and particle filters require exponentially many particles in the state dimension.
\emph{Generative Distribution Embeddings} (GDEs) learn compact representations of complex distributions but have no sequential update.
We present \emph{Neural Bayesian Filtering} (NBF), which represents beliefs as embeddings and approximates their Bayesian update.
At each step, NBF samples particles from a learned conditional generator, propagates them through the dynamics, and embeds the propagated set, weighted by the likelihood of the new observation.
Regenerating the set from the embedding at each step, rather than resampling a surviving pool, mitigates the risk of \emph{impoverishment} that degrades particle filters.
The system dynamics and likelihood enter only through propagation and weighting, so NBF retains zero-shot adaptability to new dynamics as long as the beliefs they produce fall within the family the embedding was trained on.
Training the GDE to represent that family requires only the states realized at training time.
We validate NBF on Lorenz-96 and a pursuit--evasion domain, showing graceful scaling with state dimension, accurate tracking of multimodal posteriors, and zero-shot generalization to dynamics and likelihoods held out from training.
\end{abstract}

\section{Introduction}
\label{sec:intro}
Tracking the evolution of probability distributions over time is a fundamental problem across science and engineering.
Applications range from state estimation and sequential decision-making under partial observability \citep{kaelbling1998planning} to tracking population-level responses to perturbation \citep{haug2012bayesian,barfoot2024state,sarkka2023bayesian}.
When the underlying distributions are high-dimensional, multimodal, or non-Gaussian, maintaining accurate estimates is challenging.

Classical filtering methods fall into two broad categories.
Parametric filters such as the Kalman filter \citep{kalman1960new}, its nonlinear extensions \citep{sorenson1966kalman,julier2004unscented,arasaratnam2009cubature}, and the ensemble Kalman filter \citep{evensen1994sequential} maintain a Gaussian posterior, yielding efficient updates that scale gracefully with state dimension.
However, their distributional assumptions are violated when posteriors are multimodal or non-Gaussian.
Sample-based filters such as particle filters \citep{doucet2009tutorial} can, in principle, represent arbitrary distributions through weighted particle sets, but the number of particles required for accurate approximation grows rapidly with the dimensionality of the state space.
Particle filters also suffer from \emph{impoverishment}: repeated resampling duplicates high-weight particles while discarding the rest, so the particle set progressively collapses to a few ancestral samples and loses distributional coverage \citep{sokota2022fine}.
Neither class satisfactorily handles distributions that are simultaneously complex and high-dimensional, motivating approaches that learn and maintain compact distributional representations.

Recent work on \emph{Generative Distribution Embeddings} (GDEs) shows that learned, distributionally invariant encoders coupled with conditional generators can represent complex distribution families \citep{fishman2025generative}.
In principle, these embeddings are predictive sufficient statistics, capturing all information needed for downstream tasks.
Through an associated generative model, they enable sampling (and density estimation, in some cases) on families of distributions, assuming only finite sample access to each distribution.
Critically, GDEs are static: embeddings represent single distributions, with no mechanism for sequential updates between them.

We combine classical filters and GDEs in an algorithm, which we call \emph{Neural Bayesian Filtering} (NBF), that runs Bayesian updates on particles drawn from the GDE's generator.
Given an embedding, the method (i) generates particles from the conditional generator, (ii) propagates each through the system dynamics, (iii) weights them by observation likelihood, and (iv) embeds the resulting weighted set as a representation of the posterior.
The dynamics and likelihood enter only at steps (ii) and (iii), so neither the embedding nor the generator depends on them, and NBF adapts to new dynamics without retraining, provided the beliefs those dynamics induce are ones the GDE was trained to represent.
Step (i) also draws a fresh set from the embedding at every update, so NBF never resamples and the collapse described above cannot accumulate over the steps of an episode.
The GDE is trained on the particle sets NBF itself produces, with the realized state as the only target, so training never requires samples from the beliefs themselves.

NBF is the first method to combine the representational benefits of distribution embeddings with the generalization properties of classical Bayesian filters.
Experiments on Lorenz-96 and a pursuit--evasion domain verify four properties: (1) graceful scaling with state dimension, (2) faithful representation of multimodal, non-Gaussian posteriors, (3) native parametric density estimation, and (4) zero-shot generalization to dynamics and observation likelihoods held out from training.
We also give a finite-horizon consistency result for NBF, showing that its estimates converge almost surely to the true expectation under standard particle-filter assumptions and an exact embedding.

\section{Background}
\label{sec:bg}

\subsection{Bayesian Filtering}
An environment $\env \defeq (\statespace, \obsspace, \transitionfunc, \obsfunc, \initialstatedist)$ specifies a state space $\statespace$, an observation space $\obsspace$, a transition function $\transitionfunc$, an observation function $\obsfunc$, and an initial state distribution $\initialstatedist$.
A policy $\jointpolicy(\action \mid \state)$ draws an action $\action$ from an action space $\actionspace$ in each state, and the transition function $\transitionfunc(\state^\prime \mid \state, \action)$ conditions on that action.
The filter does not observe the actions, so it marginalizes them, and we overload $\transitionfunc$ for the transition under the policy, $\transitionfunc(\state^\prime \mid \state, \jointpolicy) \defeq \int_{\actionspace} \jointpolicy(\action \mid \state)\, \transitionfunc(\state^\prime \mid \state, \action)\, d\action$.

A \emph{belief state} $\bdist$ is the posterior over the Markov state $\state \in \statespace$ given the observations so far.
Given $b_t$ and a new observation $\obs \in \obsspace$, the Bayesian update is
\begin{equation} \label{eq:bayes}
b_{t+1}(\state^\prime) \propto \int_\statespace b_t(\state) \, \kernel(\state^\prime|\state) \, d\state, \qquad \kernel(\state^\prime|\state) \defeq \transitionfunc(\state^\prime \mid \state, \jointpolicy) \, \obsfunc(\obs \mid \state, \state^\prime),
\end{equation}
where $\kernel$ is the kernel induced by the transition under $\jointpolicy$ and the observation function, and normalization is over $\state^\prime \in \statespace$.
This update is exact but requires integrating over $\statespace$, which is generally intractable when the state space is large or continuous.
Classical filters approximate it with a fixed representation of the belief and evaluate $\transitionfunc$ and $\obsfunc$ inside the update, so they apply unchanged to any system whose dynamics can be evaluated.

\textbf{Kalman filters} \citep{kalman1960new} provide closed-form updates for linear Gaussian dynamics; nonlinear variants (Extended \citep{sorenson1966kalman}, Unscented \citep{julier2004unscented}, Cubature \citep{arasaratnam2009cubature}) extend the update to mild nonlinearities, and ensemble variants \citep{evensen1994sequential,burgers1998analysis} estimate the covariance from an ensemble of states, which scales it to the high-dimensional systems of data assimilation.
Every member of the family propagates a single Gaussian.
Gaussian sum filters \citep{alspach1972nonlinear} propagate a mixture of them instead, which recovers separated modes but fits a Gaussian to each.

\textbf{Particle filters} \citep{doucet2009tutorial} approximate arbitrary distributions through weighted sample sets, but scale poorly with state dimensionality \citep{snyder2008obstacles}.
They also suffer from \emph{impoverishment}: repeated resampling concentrates the weight onto a few ancestors, so the empirical distribution loses support that the true posterior still has \citep{sokota2022fine}.
Regularization, MCMC moves, and better proposals \citep{musso2001improving,gilks2001following,doucet2009tutorial} slow the loss but keep the surviving pool, so the loss still compounds over an episode.

We consider the problem of sequentially tracking beliefs $\bdisttime$ as observations arrive.
At \emph{training time} we assume access to $\env$ and to a set of policies $\{\jointpolicy_1, \dots, \jointpolicy_J\}$, and the encoder and generator are fit to the beliefs these induce.
In a system without actions, a parameter of $\transitionfunc$ that varies between episodes takes the place of the policy, such as the forcing of Lorenz-96 in Section~\ref{sec:experiments}.
At \emph{test time} we are given $\env$ and a single policy $\jointpolicy$, and the filter evaluates $\transitionfunc(\cdot \mid \cdot, \jointpolicy)$ and $\obsfunc$ directly, so no part of the dynamics is learned.

\subsection{Distribution Embeddings}
Generative Distribution Embeddings (GDEs) provide a principled framework for representing families of distributions via learned encoders and conditional generators \citep{fishman2025generative}.
Consider a family of distributions $\mc P$ over a sample space, accessed through finite sample sets $\state_{1:n} \sim p$ for $p \in \mc P$.
An encoder $\embedfunc$ is \emph{distributionally invariant} if it depends on the sample set only through its empirical distribution, so reordering the set or duplicating every sample leaves its output unchanged.
Mean-pooled architectures (DeepSets, Set Transformers without positional encoding) satisfy this property \citep{fishman2025generative}.
A conditional generator $\generator$ takes an embedding $\bembed$ and produces samples approximating the embedded distribution.

Two properties that \citet{fishman2025generative} establish under regularity conditions make GDEs suitable for filtering.
First, distributionally invariant embeddings are asymptotically \emph{predictive sufficient statistics}: as the sample set grows, the embedding $\embedfunc(\state_{1:n})$ retains all information in the set relevant for predicting new samples from the underlying distribution.
This justifies treating it as a faithful surrogate for the beliefs when propagating dynamics and conditioning on observations.
Second, if the encoder--generator pair can represent the target distribution $p$ exactly, i.e.\ $\generator(\embedfunc(p)) = p$, then the plug-in $\generator(\embedfunc(p_n))$ formed from the empirical measure $p_n$ of $n$ i.i.d.\ samples converges weakly to $p$ as $n \rightarrow \infty$.
Together these properties say that a well-trained GDE is a consistent, lossless surrogate for the underlying distribution, with error vanishing as the provided sample set size increases.
Distribution embeddings, however, have no built-in sequential update.

\section{Embedding Belief States} \label{sec:dist}
GDEs pair a distributionally invariant encoder with a conditional generator (Section~\ref{sec:bg}).
This section instantiates one for beliefs: the encoder maps a weighted particle set representing a belief to an embedding, and the conditional generator turns that embedding into a tractable approximation of the belief.
The generator must support sampling, which the filtering update of Section~\ref{sec:filter} requires at every step, and density evaluation, which scores beliefs against the realized state on held-out data.

\paragraph{Encoder.}
Let $\bsamples \defeq \bsamplesexpanded$ be a set of particles with weights $\weightvec$ that represents a belief $\bdist$.
The encoder $\embedfunc : \bigcup_{n \geq 1} \statespace^n \times \R^n \rightarrow \R^{d_{\text{emb}}}$ is distributionally invariant and maps the weighted set to a $d_{\text{emb}}$-dimensional embedding $\bembed \defeq \embedfunc(\bsamples, \weightvec)$.
Together with the conditional generator described next, $\bembed$ defines the approximation $\bdistapprox$.

Our $\embedfunc$ is a multi-round Set Transformer: each round applies an Induced Set Attention Block (ISAB) \citep{lee2019set} followed by a pool over the particle set.
The pooled output is concatenated to every particle's input for the next round, so each round conditions on the summary of the round before it.
The weight of a particle enters the encoder alongside its state, so the embedding is a function of the weighted set.
Appendix~\ref{app:training} describes the encoders we use in Section~\ref{sec:experiments} in more detail.

\paragraph{Conditional generator.}
We use Flow Matching \citep{lipman2022flow}, which supports both sampling and density evaluation, and write $\generator(\bembed) \defeq \bdistapprox$ for the conditional generator viewed as a map from an embedding to its induced distribution.
A flow-matching model learns a vector field $v_\flowparams(z, \tau; \bembed)$, indexed by an integration time $\tau \in [0, 1]$, that defines a continuous transport from the base $\prior$ at $\tau=0$ to the target at $\tau=1$ \citep{lipman2022flow}.
Sampling draws $z \sim \prior$ and integrates the ODE $\dot{z}_\tau = v_\flowparams(z_\tau, \tau; \bembed)$ from $\tau = 0$ to $\tau = 1$, returning $\state \defeq z_1$ as the generated sample.
With the optimal-transport interpolant $z_\tau = (1-\tau)\, z + \tau\, \state$ between $z \sim \prior$ and $\state \sim \bdist$, the conditional flow-matching loss is
\[
\mathcal{L}_{\text{FM}}(\embedfuncparams, \flowparams)
= \E_{\tau, z, \state}\!\left\| v_\flowparams(z_\tau, \tau; \bembed) - (\state - z) \right\|^2.
\]
The density at $\state$ follows from integrating the divergence of $v_\flowparams$ along the same trajectory \citep{lipman2022flow}.
The encoder and generator are trained jointly on the realized states seen during training (Section~\ref{sec:training}).

\section{Filtering with Embeddings} \label{sec:filter}
Section~\ref{sec:dist} embeds a single belief as $\bembed = \embedfunc(\bsamples, \weightvec)$ and decodes it by sampling from the conditional generator.
Filtering needs an update that maps an embedding and a new observation to the embedding of the posterior, and it needs that update to hold for any dynamics the simulator can evaluate.
The second requirement rules out a purely parametric update that maps $(\bembed, \obs)$ to $\bembed^\prime$: the transition and observation models are not among its inputs, so its dependence on them is fixed at training time and the filter is tied to the dynamics it was trained on.
Classical filters meet both requirements by evaluating $\transitionfunc$ and $\obsfunc$ inside the update, and NBF does the same by running the update on particles drawn from the embedding.

A single update draws $n$ particles from $\generator(\bembed)$, propagates each through $\transitionfunc(\cdot \mid \state, \jointpolicy)$, weights each by $\obsfunc(\obs \mid \state, \state^\prime)$, and embeds the weighted set as $\bembed^\prime$.
Figure~\ref{fig:nbf-alg} sketches the loop, and Algorithm~\ref{alg:neural-filtering} gives the procedure in full.
Where a Kalman filter fits a Gaussian to the weighted set and a particle filter resamples it, NBF embeds it, and the next set is drawn fresh from $\bembed^\prime$, so no particle survives from one step to the next.
In data-assimilation terms, propagation is the forecast step, and weighting and embedding together form the analysis step, whose output is the embedding rather than an analysis ensemble.
Neither the encoder nor the generator is a function of $\transitionfunc$, $\obsfunc$, or $\jointpolicy$, so the trained model is unchanged when any of them changes.

\paragraph{Proposals.}
NBF admits the same choice of proposal as a particle filter.
Particles can be propagated through any proposal kernel $\tilde{\transitionfunc}(\state^\prime \mid \state, \obs)$ in place of the transition function, so the pairs $(\state, \state^\prime)$ are drawn from the proposal $\generator(\bembed) \cdot \tilde{\transitionfunc}$ and carry the importance weight
\[
\weightfunction(\state, \state^\prime) \defeq \frac{\transitionfunc(\state^\prime \mid \state, \jointpolicy)\, \obsfunc(\obs \mid \state, \state^\prime)}{\tilde{\transitionfunc}(\state^\prime \mid \state, \obs)}.
\]
The kernel must be positive wherever $\transitionfunc(\cdot \mid \state, \jointpolicy)\, \obsfunc(\obs \mid \state, \cdot)$ is, so that the weighted pairs represent the Bayes update of Eq.~\ref{eq:bayes}.
The bootstrap kernel $\tilde{\transitionfunc}(\cdot \mid \state, \obs) = \transitionfunc(\cdot \mid \state, \jointpolicy)$ gives $\weightfunction = \obsfunc$ and only requires sampling from $\transitionfunc(\cdot \mid \state, \jointpolicy)$; other kernels also require evaluating the transition density.

\begin{figure}[t]
\centering
\begin{tikzpicture}[
    >=Latex,
    node distance=12mm and 16mm,
    every node/.style={font=\small},
    /utils/exec=\definecolor{mplBluesA}{HTML}{F0F6FC}%
        \definecolor{mplBluesB}{HTML}{D6E6F4}%
        \definecolor{mplBluesC}{HTML}{ABD0E6}%
        \definecolor{mplObs}{HTML}{FDD9A6},
    panel/.style={
        draw, rounded corners=2pt, thick,
        minimum width=22mm, minimum height=14mm,
        align=center, inner sep=2pt,
        fill=white,
    },
    embed/.style={panel, fill=mplBluesA},
    particles/.style={panel, fill=mplBluesB},
    reweight/.style={panel, fill=mplBluesC},
    oplabel/.style={font=\scriptsize, align=center, inner sep=1pt},
    linelabel/.style={font=\scriptsize\itshape, text=gray!55!black, inner sep=1pt},
    obsnode/.style={
        draw=orange!70!black, rounded corners=1.5pt, thick,
        minimum width=8mm, minimum height=6mm,
        fill=mplObs, font=\small,
    },
    arr/.style={->, thick, shorten >=1pt, shorten <=1pt},
]

\node[embed] (theta) {%
    $\bembed$\\[1pt]%
    \scriptsize{\color{gray!50!black}embedding}%
};

\node[particles, right=of theta] (gen) {%
    \begin{tikzpicture}[baseline]
        \def\r{0.7mm}
        \fill (-0.5,  0.3) circle (\r);
        \fill (-0.2, -0.1) circle (\r);
        \fill ( 0.1,  0.4) circle (\r);
        \fill ( 0.4,  0.0) circle (\r);
        \fill (-0.4, -0.3) circle (\r);
        \fill ( 0.5, -0.4) circle (\r);
        \fill ( 0.0, -0.4) circle (\r);
        \fill ( 0.3,  0.2) circle (\r);
    \end{tikzpicture}%
};
\node[oplabel, below=0pt of gen] {\scriptsize $\particleset \iidsample \bdistapprox$};

\node[reweight, right=of gen] (rw) {%
    \begin{tikzpicture}[baseline]
        \fill (-0.45, 0.20) circle (0.4mm);
        \fill (-0.10,-0.20) circle (1.1mm);
        \fill ( 0.20, 0.35) circle (0.5mm);
        \fill ( 0.45, 0.05) circle (1.3mm);
        \fill (-0.30,-0.40) circle (0.6mm);
        \fill ( 0.50,-0.35) circle (0.9mm);
        \fill ( 0.05,-0.30) circle (1.0mm);
        \fill ( 0.35, 0.25) circle (0.5mm);
    \end{tikzpicture}%
};
\node[oplabel, below=0pt of rw] {\scriptsize $\{\particle^\prime, \particleweight\}_{i=1}^n$};

\node[embed, right=of rw] (thetap) {%
    $\bembed^\prime$\\[1pt]%
    \scriptsize{\color{gray!50!black}updated embedding}%
};

\draw[arr] (theta) -- (gen)
    node[midway, above, oplabel] {$\generator$}
    node[midway, below, linelabel, text=red!60!black] {generate};

\draw[arr] (gen) -- (rw) coordinate[midway] (propmid)
    node[midway, above=0.5pt, oplabel] {$\transitionfunc,\ \obsfunc,\ \jointpolicy$}
    node[midway, below, linelabel, text=red!60!black] {propagate};

\draw[arr] (rw) -- (thetap)
    node[midway, above, oplabel] {$\embedfunc$}
    node[midway, below, linelabel, text=red!60!black] {embed};

\node[obsnode, above=11mm of propmid] (obs) {$\obs$};
\draw[arr] (obs.south) -- ([yshift=5mm]propmid)
    node[pos=0.35, right, linelabel,  text=red!60!black] {observation};

\end{tikzpicture}
\caption{NBF update. The conditional generator draws $n$ particles from $\bembed$; each is simulated through $\transitionfunc(\cdot \mid \state, \jointpolicy)$ and reweighted by $\obsfunc(\obs \mid \state, \state^\prime)$; the weighted set is embedded into the next embedding $\bembed^\prime$.}
\label{fig:nbf-alg}
\end{figure}

\begin{algorithm}[ht]
\SetAlgoLined
\DontPrintSemicolon
\SetKwInOut{Input}{input}
\SetKwInOut{Output}{output}
\SetKwComment{Comment}{// }{}
\Input{$\bembed$ (belief embedding), $\obs$ (observation), $\transitionfunc, \obsfunc, \jointpolicy$ (system model), $\tilde{\transitionfunc}$ (proposal kernel), $n$ (number of particles), $\testfunction$ (test function)}
\Output{$\bembed^\prime$ (updated belief embedding), $\estimate$ (estimate of the posterior expectation of $\testfunction$)}
$\particleset \iidsample \generator(\bembed)$ \Comment*[r]{generate particles from the conditional generator}
\lFor{$i \leftarrow 1 \dots n$}{$\particle^\prime \sim \tilde{\transitionfunc}(\cdot \mid \particle, \obs)$, $\ \particleweight \leftarrow \weightfunction(\particle, \particle^\prime)$ \Comment*[f]{propagate and weight}}
$\estimate \leftarrow \snisfullprime$ \Comment*[r]{self-normalized estimate}
\Return $\embedfunc(\particleset^\prime, \weightvec),\ \estimate$ \Comment*[r]{embed the weighted set}
\caption{Neural Bayesian Filtering\label{alg:neural-filtering}}
\end{algorithm}

\subsection{Training} \label{sec:training}
GDEs are trained under a plug-in scheme in which each training distribution contributes a set of samples, half used to compute $\bembed$ and half to score the conditional generator \citep{fishman2025generative}.
In filtering problems those distributions are beliefs, and sampling a belief is the filtering problem itself.
NBF instead trains on the sets the filter itself embeds, which carry whatever error the model has accumulated along the episode, and scores the generator on the single realized state provided by the environment.

Training data is generated by simulating the environment and filtering with NBF under the current GDE parameters.
An episode draws an initial state\footnote{In decision processes, a policy must also be sampled from a set of training policies.} and rolls the environment forward for $\T$ steps, which yields the states $\statetime{0}, \dots, \statetime{\T}$ and the observations $\obstime{1}, \dots, \obstime{\T}$.
The filter is also given the initial belief, the policy, and any parameters of $\transitionfunc$ and $\obsfunc$ that vary between episodes.
We collect these into $\obstime{0}$ and treat them as an observation, so $b_t$ in Proposition~\ref{prop:realized-state} is the posterior of the episode process given $\obstime{0:t}$.
NBF is then run along the observations.
At each step it generates particles from the previous embedding, propagates them, and weights them by the new observation, and $\weightedparticleset_t$ denotes the resulting weighted set that Algorithm~\ref{alg:neural-filtering} embeds at step $t$, with $\weightedparticleset_0$ drawn from the initial belief.
The filter runs on the observations alone, and the realized states are used only as targets.
Particle sets for filtering are generated under the current parameters, and the loss is differentiated with the sets treated as constants, so no gradient passes back through the rollout that produced them.
Given the generated sets and realized states, the objective is
\nobreak
\begin{equation} \label{eq:realized-state-loss}
\shortloss \defeq \E\left[ \frac{1}{\T + 1} \sum_{t=0}^{\T} \ell_\flowparams\big(\statetime{t};\, \embedfunc(\weightedparticleset_t)\big) \right],
\end{equation}
\nobreak
where the expectation is over episodes and over the particle sets NBF generates along them.
$\ell_\flowparams(\state; \bembed)$ is the generator's loss at the single target $\state$ given the embedding $\bembed$.

An episode draws $\statetime{t}$ and $\obstime{0:t}$ from the same process that defines $b_t$ ($\statetime{t} \sim b_t$, given the observations) and $\weightedparticleset_t$ is a function of those observations and the filter's randomness.
When $\ell_\flowparams$ is the negative log-density, the objective is the expected cross-entropy of the true posterior under the generator's belief.
It is a strictly proper scoring rule \citep{gneiting2007strictly}, so its unique minimizer over conditional densities of the state given the set is the \textbf{set posterior} $p(\cdot \mid \weightedparticleset_t)$.
Every target is a realized state, so the objective needs no reference posterior and the generator is never scored against its own samples.
Proposition~\ref{prop:realized-state} decomposes the loss into three terms (proof in Appendix~\ref{app:proofs}).

\begin{proposition}[Realized-state objective] \label{prop:realized-state}
Let $\weightedparticleset_t$ be a random set that is a function of $\obstime{0:t}$ and of randomness independent of the episode, and assume that $\statetime{t}$ has finite entropy, both unconditionally and given $\obstime{0:t}$.
For any conditional density $\rho(\cdot \mid \weightedparticleset)$ over states given a set,
\[
\E\big[-\log \rho(\statetime{t} \mid \weightedparticleset_t)\big]
= \mathrm{H}\big(\statetime{t} \mid \obstime{0:t}\big)
+ \E\big[\mathrm{KL}\big(b_t \,\|\, p(\cdot \mid \weightedparticleset_t)\big)\big]
+ \E\big[\mathrm{KL}\big(p(\cdot \mid \weightedparticleset_t) \,\|\, \rho(\cdot \mid \weightedparticleset_t)\big)\big].
\]
\end{proposition}

The GDE parameters that generate an episode's sets are fixed during the episode, so each set is a function of the episode's observations, and this decomposition applies throughout training.
The first term is the entropy of the belief, which no filter can score below.
The second is the information about the state that the set has lost relative to the observations; it depends on the sets and not on $\rho$, and it is zero when the set determines $b_t$.
The third is the model error, the only term that depends on $\rho$, so a trained generator's excess loss is its expected KL to the set posterior.

For a generator with a tractable likelihood, $\ell_\flowparams(\state; \bembed)$ is the negative log-density of $\state$ under $\generator(\bembed)$.
For flow matching it is $\mathcal{L}_{\text{FM}}$ with $\state$ as the only target, which has the same minimizer over conditional distributions \citep{lipman2022flow}.
Gradients flow through the embedding of each set and through the generator, and particle sets are never stored across transitions.

\subsection{Convergence of NBF} \label{sec:convergence}
NBF is \emph{consistent} under the following assumptions:
\begin{enumerate}
    \item[(A1)] \emph{Polish state and observation spaces.} $\statespace$ and $\obsspace$ are separable and completely metrizable.
    \item[(A2)] \emph{GDE regularity.} $\embedfunc$ is distributionally invariant \citep{fishman2025generative} in the weighted sense: it depends on a weighted set only through the set's weighted empirical measure, and it extends to probability measures on $\statespace$.
    The induced plug-in map $\Phi(P) \defeq \generator(\embedfunc(P))$ is well-specified ($\Phi(b_t) = b_t$) and weakly continuous at every belief state $b_t$, $t \leq \T$, along the observation sequence.
    \item[(A3)] \emph{Regular weight and proposal kernel.} At each step $t < \T$, let $\proposal$ denote the proposal distribution over pairs $(\state, \state^\prime)$ obtained by drawing $\state \sim b_t$ and then $\state^\prime \sim \tilde{\transitionfunc}(\cdot \mid \state, \obstime{t+1})$, and let $\weightfunction(\state, \state^\prime)$ be the importance weight for the observation $\obstime{t+1}$.
    We require:
    \begin{enumerate}
        \item[(i)] \emph{Bounded continuous weight:} $\weightfunction$ is continuous and takes values in $[0, W_{\max}]$ for some $W_{\max} < \infty$.
        \item[(ii)] \emph{Non-degenerate likelihood:} $\E_\proposal[\weightfunction] > 0$.
        \item[(iii)] \emph{Feller kernel \citep{crisan2002survey}:} if $\state_k \to \state$ then $\tilde{\transitionfunc}(\cdot \mid \state_k, \obstime{t+1})$ converges weakly to $\tilde{\transitionfunc}(\cdot \mid \state, \obstime{t+1})$.
    \end{enumerate}
\end{enumerate}
Assumptions (A1) and (A3) are the standard conditions of particle filter convergence theory \citep{crisan2002survey}.
Assumption (A2) is an idealization of the trained model.
It requires the generator to return every belief the filter reaches from that belief's embedding, and a weighted set whose empirical measure is close to that belief to decode to a distribution close to it, so the sampling error of the particle set is the only error left in the update.

\begin{theorem}[NBF Consistency] \label{thm:nbf-consistency}
Fix a finite horizon $\T$ and an observation sequence $\obstime{1:\T}$ with belief states $b_t$, $t \leq \T$, for which Assumptions (A1)--(A3) hold, and let $\testfunction:\statespace \rightarrow \R$ be bounded and continuous.
For $t \geq 1$, let $\testimate$ be the self-normalized estimate of $\ttargetexpectation$ computed by NBF using Algorithm~\ref{alg:neural-filtering} with $n$ particles per step, and let $\hat\mu^{(n)}_0(\testfunction)$ be the average of $\testfunction$ over the $n$ particles drawn from $\initialstatedist$.
Then $\maxabsdiff \asconv 0$ as $n \rightarrow \infty$.
\end{theorem}

We add one further condition, on the error the embedding adds at each step, to establish the standard Monte-Carlo rate.
Further details, including proofs, are included in Appendix~\ref{app:proofs}.

\begin{corollary}[NBF Convergence Rate] \label{cor:nbf-rate}
Under the same conditions as Theorem~\ref{thm:nbf-consistency}, suppose additionally that the embedding error is of Monte-Carlo order, $\E_{\generator(\bembed_t)}[g] - \hat\mu^{(n)}_t(g) = O_p(n^{-1/2})$ for every $t \leq \T$ and every bounded continuous $g$, where $\bembed_t$ is the embedding at step $t$ and $\hat\mu^{(n)}_t$ is the weighted empirical measure of the set it embeds.
Then $\maxabsdiff = O_p(n^{-1/2})$.
\end{corollary}

\section{Experiments} \label{sec:experiments}
We evaluate NBF on Lorenz-96, a chaotic system in up to 40 dimensions whose posterior is unimodal but non-Gaussian, and on pursuit--evasion, a tracking problem whose posterior is multimodal.
The experiments test four properties: representation of multimodal posteriors, graceful scaling with state dimension, native density estimation, and zero-shot generalization to dynamics and likelihoods held out from training.
More details about both domains can be found in Appendix~\ref{app:domains}.

\paragraph{Lorenz-96.}
The state is a ring of $K \in \{20, 40\}$ coordinates under the chaotic dynamics of \citet{lorenz1996predictability} with forcing $F$.
Every second site is observed, perturbed with Gaussian noise, once per window of 0.4 time units.
This time interval is long enough that the forecast is nonlinear between observations (Appendix~\ref{app:l96-domain}).
The forcing $F$ is the quantity that varies between episodes (Section~\ref{sec:bg}).
Training draws $F$ from $[5, 7] \cup [9, 11]$ and $\sigma_{\text{obs}}$ from $[0.5, 1.1]$.
Evaluation runs at $F = 6$ and $F = 10$ inside the training range and at the held-out $F = 8$, at $\sigma_{\text{obs}} = 1$ unless stated.
The likelihood factors over the $K/2$ observed sites, and the product of that many factors concentrates the weight of a particle set on a few particles from the first update \citep{snyder2008obstacles}.
On this domain the particle weights provided to NBF's encoder are the vectors of standardized residuals $(\obs_j - \state^\prime_j)/\sigma_{\text{obs}}$ over the observed sites, with zero at unobserved sites.

\paragraph{Pursuit--evasion.}
The agent tracks $k$ evaders moving through an obstacle-filled arena with a single stationary range sensor. 
Each evader follows a stochastic heading policy drawn from a training mixture of four, and four further policies are held out for evaluation (Appendix~\ref{app:policies}).
The range noise $\sigma$ is drawn once per episode from $[0.01, 0.05]$.
Both evader movements and sensor range readings are perturbed with Gaussian noise.
The $k$ range readings output by the sensor are randomly shuffled before they reach the filter, so the likelihood is a sum over the $k!$ assignments of ranges to evaders.
This likelihood does not factor over the joint state in $\R^{2k}$, and induces a multimodal posterior.

\paragraph{Methods.}
Every filter is given the same dynamics, observation model, and initial belief.
\textbf{NBF$(n)$} is our method with $n$ particles in the generate-propagate-embed loop.
\textbf{PF$(n)$} is a bootstrap (SIR) particle filter \citep{gordon1993novel} with $n$ particles and systematic resampling, and \textbf{PF-Reg$(n)$} is a regularized particle filter \citep{musso2001improving}, which jitters the resampled set with the shrinkage kernel of \citet{liu2001combined} to counter impoverishment.
\textbf{EnKF$(m)$} is a stochastic ensemble Kalman filter with $m$ members \citep{evensen1994sequential,burgers1998analysis}, and \textbf{GMF$(m)$} is a Gaussian sum filter with $m$ components \citep{alspach1972nonlinear}, run on pursuit--evasion only.
Results for resample-move and locally optimal proposal particle filters appear in Appendix~\ref{app:add-experiments}, and the proposals and settings of every baseline are detailed in Appendix~\ref{app:baselines}.

\paragraph{Metrics.}
Every method is scored against the realized state.
NLL is the negative log-density of the true state under the filter's density.
Particle filters have no native density, so we also measure the continuous ranked probability score (CRPS) \citep{gneiting2007probabilistic} of each marginal against the true coordinate.
CRPS is computed from an equal-weight sample for every filter.
RMSE of the posterior mean is the standard data-assimilation score and is reported on Lorenz-96.
We do not report RMSE on pursuit--evasion, since the mean of a multimodal belief can lie between its modes, where the posterior has little mass.
Scores are averaged over the steps of an episode after a five-step spin-up and over 500 episodes, generated from five evaluation seeds shared by every method.
Every NBF entry is the mean over three independently trained models with the SEM over those models.
For other filters with no training phase, SEM is computed over episodes.
The tables in this section report the headline cells, and Appendix~\ref{app:add-experiments} reports every method at every particle budget with additional calibration metrics.

\begin{table}[t]
    \centering
    \setlength{\tabcolsep}{3pt}
    \caption{Lorenz-96 at the held-out $F = 8$. \emph{ms} is wall-clock per filter step on one L40S GPU. SEMs are at most 0.2 nats on NLL and 0.03 on RMSE and CRPS.}
    \label{tab:l96-main}
    \footnotesize
    \begin{tabular}{lrrrrrrrr}
\toprule
 & \multicolumn{4}{c}{$K = 20$} & \multicolumn{4}{c}{$K = 40$} \\
\cmidrule(lr){2-5} \cmidrule(lr){6-9}
Method & NLL & RMSE & CRPS & ms & NLL & RMSE & CRPS & ms \\
\midrule
NBF(1024) (ours) & 23.4 & 0.965 & 0.524 & 3.6 & 53.6 & 1.060 & 0.564 & 4.0 \\
NBF(4096) (ours) & \textbf{23.1} & 0.943 & 0.513 & 11.0 & \textbf{52.6} & \textbf{1.029} & \textbf{0.549} & 11.8 \\
PF-Reg(131072) & -- & \textbf{0.877} & \textbf{0.489} & 9.4 & -- & 4.332 & 3.286 & 20.9 \\
EnKF(100) & 31.7 & 1.484 & 0.770 & 0.01 & 63.1 & 1.506 & 0.764 & 0.01 \\
\bottomrule
\end{tabular}

\end{table}

\paragraph{Scaling with state dimension.}
A particle filter's effective sample size collapses as the state dimension grows, and no practical budget, proposal, or jitter after resampling restores it.
NBF regenerates its set from its embedding at every step, so its particle budget does not have to grow as rapidly with state dimension.
NBF's error should therefore move with the EnKF's as the dimension grows, while the particle filters go from the most accurate filter at lower dimensions to collapsed at the highest.
Tables~\ref{tab:l96-main} and~\ref{tab:pe-main} show this on both domains.
The particle filters are the most accurate at $K = 20$ and $k = 1$ and the least accurate at $K = 40$ and $k \geq 3$, while NBF's error follows the EnKF's and stays below it on Lorenz-96.
Quadrupling NBF's particles barely changes its error, as expected due to its cardinality-invariant encoder.

\begin{table}[t]
    \centering
    \setlength{\tabcolsep}{3pt}
    \caption{Pursuit--evasion in distribution, averaged with equal weight over the four policies of the training mixture. \emph{ms} is as in Table~\ref{tab:l96-main}. SEMs are at most 1.0 nats on NLL for NBF and 2.6 for the EnKF, and 0.03 on CRPS for all methods.}
    \label{tab:pe-main}
    \footnotesize
    \begin{tabular}{lrrrrrrrrr}
\toprule
 & \multicolumn{3}{c}{$k = 1$} & \multicolumn{3}{c}{$k = 3$} & \multicolumn{3}{c}{$k = 5$} \\
\cmidrule(lr){2-4} \cmidrule(lr){5-7} \cmidrule(lr){8-10}
Method & NLL & CRPS & ms & NLL & CRPS & ms & NLL & CRPS & ms \\
\midrule
NBF(512) (ours) & -0.01 & 1.285 & 0.44 & 11.6 & 1.087 & 1.6 & 22.0 & 0.937 & 2.2 \\
NBF(2048) (ours) & \textbf{-0.07} & 1.280 & 1.6 & \textbf{10.5} & 1.070 & 4.5 & \textbf{20.3} & 0.926 & 5.8 \\
PF(32768) & -- & \textbf{1.090} & 0.071 & -- & 1.698 & 0.23 & -- & 1.685 & 0.49 \\
PF-Reg(32768) & -- & 1.157 & 0.091 & -- & 1.678 & 0.28 & -- & 1.698 & 0.61 \\
EnKF(100) & 5.25 & 1.500 & 0.008 & 15.0 & \textbf{1.025} & 0.008 & 31.1 & \textbf{0.897} & 0.009 \\
GMF(64) & 26.8 & 1.697 & 0.12 & 958.6 & 1.320 & 0.18 & 3703.4 & 1.199 & 0.30 \\
\bottomrule
\end{tabular}

\end{table}

\paragraph{Multimodal posteriors.}
In pursuit--evasion, a single range constrains an evader to a ring around the sensor, so the belief is annular at $k = 1$ and multimodal at every $k$.
A single Gaussian centred on the ring places its mass inside and around it, where the true state is not, so its density at the true state should be low however accurate its mean, while NBF's generator can place the mass directly on the ring.
In Table~\ref{tab:pe-main}, NBF has the lowest NLL at every $k$, and GMF, which fits its components to slivers of the ring, has the highest by a wide margin.
On CRPS the EnKF is worse than NBF at $k = 1$ and slightly better at $k \geq 3$, so its mean is a competitive point estimate while its Gaussian misplaces the mass.

\begin{figure}[t]
    \centering
    \includegraphics[width=\linewidth]{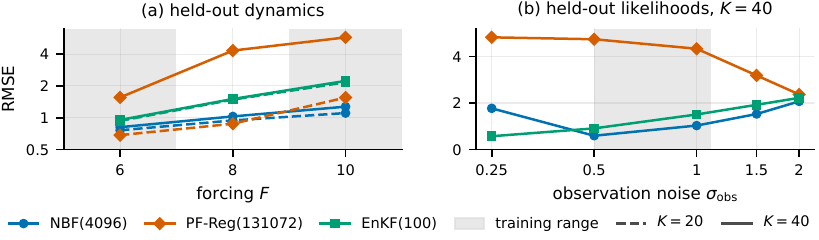}
    \caption{Zero-shot generalization on Lorenz-96. (a) RMSE against the forcing $F$ at $K = 20$ and $K = 40$. (b) RMSE against the observation noise at $K = 40$. Shading marks the training ranges, and SEMs are below 0.1.}
    \label{fig:zero-shot-l96}
\end{figure}

\begin{figure}[t]
    \centering
    \includegraphics[width=\linewidth]{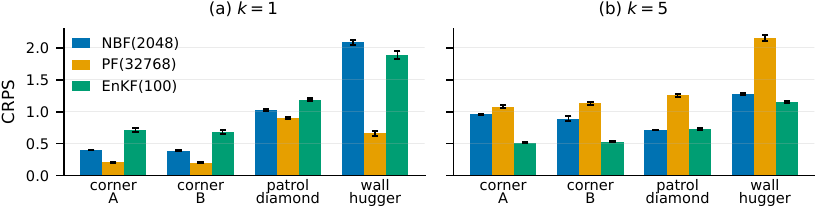}
    \caption{Zero-shot generalization on four held-out policies in pursuit--evasion. Error bars are SEMs over trained models for NBF and over episodes for the other filters.}
    \label{fig:zero-shot-pe}
\end{figure}

\paragraph{Zero-shot generalization.}
System dynamics enter NBF only through propagation, so the algorithm should generalize to a held-out forcing or policy unless the beliefs it induces fall outside the family the encoder was trained on.
Likewise, the likelihood enters through weighting, and the encoder is given the weights but not the likelihood's parameters, so a held-out likelihood should cost nothing where the weights it produces resemble those seen in training but could fail where they do not.
Figure~\ref{fig:zero-shot-l96} plots RMSE on Lorenz-96 against forcing and observation noise.
At both $K$, NBF's RMSE at held-out $F = 8$ lies on the trend over the two training intervals, and its margin over the EnKF grows with $F$, which we attribute to the forecast ensemble moving further from Gaussian as the dynamics become more chaotic.
NBF is the most accurate filter from the bottom of the training range of $\sigma_{\text{obs}}$ to well above its top, but below it NBF trails the EnKF, since the particle weights there are unlike any the encoder saw in training.

Figure~\ref{fig:zero-shot-pe} plots CRPS on the four held-out pursuit--evasion policies.
At $k = 1$ the state is two-dimensional and the likelihood is a single factor, so PF is close to the best attainable CRPS and has the lowest CRPS on every held-out policy.
NBF trails it on all four, and by the most on \emph{wall hugger}.
We attribute the \emph{wall hugger} gap to coverage of the training distribution.
The beliefs induced by wall-following concentrate along structures that the training policies rarely visit, so the embedding has not learned to represent them.
At $k = 5$, PF trails NBF on every held-out policy, including \emph{wall hugger}, so NBF's advantage over the particle filters at high dimension can hold even on out-of-distribution policies for which it performs relatively poorly in low dimensions.

\section{Limitations} \label{sec:limitations}
NBF generalizes zero-shot to held-out dynamics and likelihoods, but only to those whose beliefs and particle weights resemble the ones seen in training.
In our experiments this limit appears below the training range of $\sigma_{\text{obs}}$ on Lorenz-96, where the particle weights are unlike any the encoder saw in training, and under the wall-following policy on pursuit--evasion, which induces beliefs that the training policies rarely visit (Figures~\ref{fig:zero-shot-l96} and~\ref{fig:zero-shot-pe}).
Widening the training range or the policy mixture to cover these cases requires no change to the update.
NBF also costs three orders of magnitude more per step than an ensemble Kalman filter, which is slightly more accurate on pursuit--evasion beliefs that collapse toward a point, so the expense is warranted when the belief is multimodal or the dimension is high enough that particle filters collapse.

Theorem~\ref{thm:nbf-consistency} assumes in (A2) an embedding that represents every reachable belief exactly, so it establishes the correctness of the update in isolation and bounds none of the representation error of a trained encoder, which is of fixed size and does not necessarily improve with $n$.
Proposition~\ref{prop:realized-state} identifies the information the set has lost without bounding the error, so the distance from the set posterior to $b_t$ at finite $n$ is measured only empirically.
We leave a bound on that term to future work.
Both results concern the update and the objective, so a better encoder or generator improves the filter without any change to either.

NBF assumes that $\transitionfunc$ and $\obsfunc$ are given as input to the problem, since every update evaluates them.
It therefore cannot be applied as-is to a system whose dynamics or observation model are unknown.
Recurrent state-space models \citep{krishnan2015deep,karl2016deep,chung2015recurrent} and KalmanNet \citep{revach2022kalmannet} learn the update itself, so their dynamics are fixed at training time, and differentiable particle filters \citep{jonschkowski2018differentiable,karkus2018particle,chen2024normalizing} learn the transition and observation models inside a particle update.
The same dynamics enter NBF only through propagation and weighting, so those learned models can replace those defined by the system without changing the update.

\section{Conclusion}
\label{sec:concl}
Neural Bayesian Filtering stores a belief as a learned embedding and updates it through a particle set.
Its conditional generator samples particles, the given dynamics and observation likelihood propagate and weight them, and its encoder embeds the result.
The embedding represents multimodal, high-dimensional beliefs, and regenerating particles at every step mitigates impoverishment.
System dynamics and likelihood enter only through propagation and weighting, so NBF adapts to dynamics unseen at training time as long as the beliefs they induce lie in the family the embedding was trained on.
The model is trained on the particle sets the filter itself produces and scored against the realized state, so training needs episodes with recorded single ground-truth states rather than samples from the target beliefs.

NBF represents a belief as a fixed-size, differentiable summary of the posterior, so a planner can condition on its embedding directly.
This could mitigate the particle depletion that depth-limited Monte Carlo planners \citep{silver2010monte,vsustr2021particle} observe in large games.
The generator also gives the belief a tractable density, so a planner can reward reductions in the belief's entropy without fitting a kernel density to particles \citep{fischer2020information}.
Beyond planning, Lorenz-96 and pursuit--evasion are small instances of data assimilation for chaotic geophysical models \citep{evensen1994sequential,hunt2007efficient} and of target tracking \citep{haug2012bayesian} with unlabeled measurements.
In both, the dynamics can be simulated, which NBF's training requires, and the posterior can be too high-dimensional or multimodal for Gaussian and particle filters.

\subsection*{Societal impact}
NBF is a general-purpose filter that we evaluate only on simulated systems, so its societal impact is that of state estimation in general.
Like any tracking method, it has broader potential applications in surveillance.

\subsection*{Acknowledgments}
This work was supported in part by the Canada CIFAR AI Chairs program and the Alberta Machine Intelligence Institute (Amii) and the Natural Sciences and Engineering Research Council of Canada (NSERC), and was enabled in part by support provided by Prairies DRI and the Digital Research Alliance of Canada (\url{https://www.alliancecan.ca}).
R. Haluška and M. Schmid were supported by the grant no. 25-18031S of the Czech Science Foundation (GAČR) and the Charles University Grant Agency (GAUK) project no. 326525. Additional computational resources were provided by the e-INFRA CZ project (ID:90254), supported by the Ministry of Education, Youth and Sports of the Czech Republic.

\bibliography{refs}
\bibliographystyle{plainnat}

\newpage
\appendix
\section{Proofs}
\label{app:proofs}
\begin{propositionrep}[Realized-state objective]
Let $\weightedparticleset_t$ be a random set that is a function of $\obstime{0:t}$ and of randomness independent of the episode, and assume that $\statetime{t}$ has finite entropy, both unconditionally and given $\obstime{0:t}$.
For any conditional density $\rho(\cdot \mid \weightedparticleset)$ over states given a set,
\[
\E\big[-\log \rho(\statetime{t} \mid \weightedparticleset_t)\big]
= \mathrm{H}\big(\statetime{t} \mid \obstime{0:t}\big)
+ \E\big[\mathrm{KL}\big(b_t \,\|\, p(\cdot \mid \weightedparticleset_t)\big)\big]
+ \E\big[\mathrm{KL}\big(p(\cdot \mid \weightedparticleset_t) \,\|\, \rho(\cdot \mid \weightedparticleset_t)\big)\big].
\]

\begin{proof}
Write $\weightedparticleset \defeq \weightedparticleset_t$, $\state \defeq \statetime{t}$, and $\obs \defeq \obstime{0:t}$.
Adding and subtracting $\log p(\state \mid \weightedparticleset)$ inside the expectation gives
\[
\E\big[-\log \rho(\state \mid \weightedparticleset)\big]
= \mathrm{H}(\state \mid \weightedparticleset)
+ \E\big[\mathrm{KL}\big(p(\cdot \mid \weightedparticleset) \,\|\, \rho(\cdot \mid \weightedparticleset)\big)\big].
\]
By hypothesis the set is a function of the observations and of randomness independent of the episode, so the state and the set are conditionally independent given the observations, $p(\state \mid \obs, \weightedparticleset) = p(\state \mid \obs) = b_t$, and $\mathrm{H}(\state \mid \obs, \weightedparticleset) = \mathrm{H}(\state \mid \obs)$.
By the definition of conditional mutual information,
\[
\mathrm{H}(\state \mid \weightedparticleset)
= \mathrm{H}(\state \mid \obs, \weightedparticleset) + \mathrm{I}(\state; \obs \mid \weightedparticleset)
= \mathrm{H}(\state \mid \obs) + \mathrm{I}(\state; \obs \mid \weightedparticleset),
\]
and all three quantities are finite because $\mathrm{H}(\state \mid \obs) \leq \mathrm{H}(\state \mid \weightedparticleset) \leq \mathrm{H}(\state)$.
Finally,
\[
\mathrm{I}(\state; \obs \mid \weightedparticleset)
= \E\big[\mathrm{KL}\big(p(\cdot \mid \obs, \weightedparticleset) \,\|\, p(\cdot \mid \weightedparticleset)\big)\big]
= \E\big[\mathrm{KL}\big(b_t \,\|\, p(\cdot \mid \weightedparticleset)\big)\big].
\]
\end{proof}
\end{propositionrep}

The second term $\mathrm{I}(\statetime{t}; \obstime{0:t} \mid \weightedparticleset_t)$ is the information about the state that the observations carry beyond the set.
When the set determines the current observation, that information is bounded by what the past observations add to the current one, so the set posterior scores at least as well as the posterior given the current observation alone.
The decomposition holds at every particle count and for any quality of the sets, and the set posterior is calibrated at every $n$ because it is a conditional distribution of the state given the set.
The plug-in generator $\generator(\embedfunc(\weightedparticleset_t))$ of the GDE training scheme \citep{fishman2025generative} is itself a conditional density given the set, so its expected loss is at least that of the set posterior.

\begin{lemma}[Set that determines the observation] \label{lem:set-determines-obs}
Under the hypotheses of Proposition~\ref{prop:realized-state}, if $\obstime{t}$ is a function of $\weightedparticleset_t$, then
\[
\mathrm{I}\big(\statetime{t};\, \obstime{0:t} \mid \weightedparticleset_t\big) \leq \mathrm{I}\big(\statetime{t};\, \obstime{0:t-1} \mid \obstime{t}\big),
\]
so the expected loss of the set posterior is at most $\mathrm{H}(\statetime{t} \mid \obstime{t})$.
\begin{proof}
Write $\weightedparticleset$, $\state$, and $\obs$ as above.
Conditional independence gives $\mathrm{I}(\state; \obs, \weightedparticleset) = \mathrm{I}(\state; \obs)$, so $\mathrm{I}(\state; \obs \mid \weightedparticleset) = \mathrm{I}(\state; \obs) - \mathrm{I}(\state; \weightedparticleset)$.
Expanding $\mathrm{I}(\state; \weightedparticleset, \obstime{t})$ in both orders gives $\mathrm{I}(\state; \weightedparticleset) + \mathrm{I}(\state; \obstime{t} \mid \weightedparticleset) = \mathrm{I}(\state; \obstime{t}) + \mathrm{I}(\state; \weightedparticleset \mid \obstime{t})$.
Since $\obstime{t}$ is a function of $\weightedparticleset$, $\mathrm{I}(\state; \obstime{t} \mid \weightedparticleset) = 0$, so $\mathrm{I}(\state; \weightedparticleset) \geq \mathrm{I}(\state; \obstime{t})$, and by the chain rule $\mathrm{I}(\state; \obs) - \mathrm{I}(\state; \obstime{t}) = \mathrm{I}(\state; \obstime{0:t-1} \mid \obstime{t})$.
Adding $\mathrm{H}(\state \mid \obs)$ to both sides of the bound gives $\mathrm{H}(\state \mid \obs) + \mathrm{I}(\state; \obs \mid \weightedparticleset) \leq \mathrm{H}(\state \mid \obstime{t})$, which by Proposition~\ref{prop:realized-state} is the expected loss of the set posterior.
\end{proof}
\end{lemma}

We next state the strong law for self-normalized importance sampling, which we use to handle the propagate-and-reweight step in the inductive argument.
In the filter, the particles at step $t+1$ are drawn from the generator conditioned on the embedding of the step-$t$ set, which was itself computed from $n$ random particles.
The proposal is therefore a random distribution that differs for every $n$, and the particles are i.i.d.\ only conditionally on it.
The identities follow \citet{owen2013monte}.

\begin{lemma}[Strong Law for Self-Normalized Importance Sampling Estimators] \label{lem:sllnis}
Let $\mc U$ be a Polish space, let $\weightfunction : \mc U \rightarrow [0, W_{\max}]$ be a bounded measurable weight function, and let $\testfunction: \mc U \rightarrow \R$ be any bounded measurable function.
For each $n$, let $\proposal_n$ be a random proposal distribution on $\mc U$, draw $u_1,\dots,u_n \iidsample \proposal_n$ conditionally on $\proposal_n$, and let $\particleweight = \weightfunction(u_i)$.
\begin{enumerate}
\item[(a)] $\frac{1}{n}\sum_{i=1}^n \particleweight \testfunction(u_i) - \E_{\proposal_n}[\weightfunction\testfunction] \asconv 0$ as $n \rightarrow \infty$.
\item[(b)] Let $b^{(n)} \defeq c_n \cdot \weightfunction \proposal_n$ be the target with normalization constant $c_n$, and let $\estimate = \sum_{i} \particleweight \testfunction(u_i) / \sum_{i} \particleweight$. If $\liminf_n \E_{\proposal_n}[\weightfunction] > 0$ almost surely, then $\estimate - \E_{b^{(n)}}[\testfunction] \asconv 0$.
\end{enumerate}
In particular, for a fixed proposal $\proposal$ with $\E_\proposal[\weightfunction] > 0$ and target $b = c \cdot \weightfunction \proposal$, $\estimate \asconv \E_b[\testfunction]$.
\begin{proof}
Conditionally on $\proposal_n$, the summands $\particleweight\testfunction(u_i)$ are i.i.d.\ and take values in an interval of length $2 W_{\max} \inftynorm{\testfunction}$, so Hoeffding's inequality \citep{hoeffding1963probability} gives, for any $\epsilon > 0$,
\[
\Pr\!\left( \Big| \tfrac{1}{n}\textstyle\sum_{i=1}^n \particleweight\testfunction(u_i) - \E_{\proposal_n}[\weightfunction\testfunction] \Big| > \epsilon \right) \leq 2\exp\!\left( -\frac{n \epsilon^2}{2 W_{\max}^2 \inftynorm{\testfunction}^2} \right).
\]
The bound does not depend on $\proposal_n$, so it holds unconditionally, and it is summable over $n$.
By the Borel--Cantelli lemma, the difference exceeds $\epsilon$ for only finitely many $n$ almost surely, and taking $\epsilon = 1/k$ for $k = 1, 2, \dots$ gives (a).

For (b), $b^{(n)}$ integrates to one, so $c_n^{-1} = \E_{\proposal_n}[\weightfunction]$, and multiplying and dividing by $c_n$ gives $\E_{\proposal_n}[\weightfunction\testfunction] = c_n^{-1} \bar\testfunction_n$ with $\bar\testfunction_n \defeq \E_{b^{(n)}}[\testfunction]$.
Let $N_n \defeq \sum_{i=1}^n \particleweight\testfunction(u_i)$ and $\denominator \defeq \sum_{i=1}^n \particleweight$, so that $\estimate = N_n / \denominator$, and let
\[
\alpha_n \defeq \frac{N_n}{n} - c_n^{-1} \bar\testfunction_n, \qquad \beta_n \defeq \frac{\denominator}{n} - c_n^{-1}
\]
be the errors of the numerator and of the denominator.
Part (a) applied to $\testfunction$ and to the constant function one gives $\alpha_n \asconv 0$ and $\beta_n \asconv 0$.
Substituting $N_n / n = c_n^{-1} \bar\testfunction_n + \alpha_n$ and $c_n^{-1} = \denominator / n - \beta_n$,
\[
\estimate - \bar\testfunction_n = \frac{c_n^{-1} \bar\testfunction_n + \alpha_n}{\denominator / n} - \bar\testfunction_n = \frac{\alpha_n - \bar\testfunction_n \beta_n}{\denominator / n}.
\]
By hypothesis there is almost surely a $\delta > 0$ with $c_n^{-1} \geq \delta$ for all large $n$, so $\denominator / n = c_n^{-1} + \beta_n \geq \delta / 2$ for all large $n$.
Since $|\bar\testfunction_n| \leq \inftynorm{\testfunction}$,
\[
\big|\estimate - \bar\testfunction_n\big| \leq \frac{2}{\delta}\Big(|\alpha_n| + \inftynorm{\testfunction}\, |\beta_n|\Big) \asconv 0.
\]
\end{proof}
\end{lemma}

In the filter, $\mc U = \statespace \times \statespace$, a point is a pair $u = (\state, \state^\prime)$, the weight is $\weightfunction(\state, \state^\prime)$, and test functions depend on $\state^\prime$ only.
Applying the lemma to one update of NBF gives the inductive step of the theorem below: if the distribution the generator samples from at step $t$ converges weakly to $b_t$, then the distribution at step $t+1$ converges weakly to $b_{t+1}$.

\begin{theoremrep}[NBF Consistency]
Fix a finite horizon $\T$ and an observation sequence $\obstime{1:\T}$ with belief states $b_t$, $t \leq \T$, for which Assumptions (A1)--(A3) hold, and let $\testfunction:\statespace \rightarrow \R$ be bounded and continuous.
For $t \geq 1$, let
\[
\testimate = \snisfullprime
\]
be the estimate of $\ttargetexpectation$ computed by NBF using Algorithm~\ref{alg:neural-filtering} with $n$ particles per step, and let $\hat\mu^{(n)}_0(\testfunction)$ be the average of $\testfunction$ over the $n$ particles drawn from $\initialstatedist$.
Then,
\[
\maxabsdiff \asconv 0
\]
as $n \rightarrow \infty$.

\begin{proof}
According to Algorithm~\ref{alg:neural-filtering}, the weight for transitioning from $\particle$ to $\particle^\prime$ is $\particleweight = \weightfunction(\particle, \particle^\prime)$.
Let $\bembed_t$ denote the embedding at step $t$ and $b_{\bembed_t} = \generator(\bembed_t)$ the distribution the generator samples from, which is random and depends on $n$.
We prove by induction that $b_{\bembed_t}$ converges weakly to $b_t$ almost surely for every $t \leq \T$, and convergence of the estimate follows.
Here weak convergence refers to distributions over $\statespace$ and almost surely to the filter's draws: with probability one over those draws, the sequence of distributions $b_{\bembed_t}$ indexed by $n$ converges weakly.

\paragraph{Base case. $t=0$}
Here $b_t = \initialstatedist$.
By Lemma~\ref{lem:sllnis}(a) with unit weights, applied to each function in a countable convergence-determining class \citep[proof of Theorem~11.4.1]{dudley2002real}, the empirical measure $\hat\mu^{(n)}_0$ of the $n$ particles drawn from $\initialstatedist$ converges weakly to $\initialstatedist$ almost surely as $n \to \infty$.
By (A2), the plug-in map is weakly continuous and well-specified at $\initialstatedist$, so $b_{\bembed_0} \defeq \generator(\embedfunc(\hat\mu^{(n)}_0)) \Rightarrow \initialstatedist$ almost surely.

\paragraph{Inductive step.}
Assume that $b_{\bembed_t}$ converges weakly to $b_t$ almost surely as $n \to \infty$.
Throughout the step, $\obs \defeq \obstime{t+1}$ is the new observation.

After the loop in Algorithm~\ref{alg:neural-filtering}, the pairs $(\particle, \particle^\prime)$ are, conditionally on $b_{\bembed_t}$, i.i.d.\ draws from $\hat\proposal(\state, \state^\prime) \defeq b_{\bembed_t}(\state)\, \tilde{\transitionfunc}(\state^\prime \mid \state, \obs)$ with weight function $\weightfunction(\state, \state^\prime)$.
The exact posterior at time $t+1$ can be written as
\begin{equation} \label{eq:p_expanded}
b_{t+1}(\state^\prime) = c \int \bdisttime\, \tilde{\transitionfunc}(\state^\prime \mid \state, \obs)\, \weightfunction(\state, \state^\prime)\, d\state
\end{equation}
for some normalization constant $c$, which is the Bayes update of Eq.~\ref{eq:bayes} with the kernel $\kernel$ multiplied and divided by $\tilde{\transitionfunc}$, since $\tilde{\transitionfunc}\, \weightfunction = \transitionfunc(\cdot \mid \cdot, \jointpolicy)\, \obsfunc$.
The integrand is the proposal $\proposal = b_t \cdot \tilde{\transitionfunc}$ over pairs from (A3) times the weight, so $b_{t+1}$ is the $\state^\prime$-marginal of the target $c\, \weightfunction \proposal$ of Lemma~\ref{lem:sllnis} with $c^{-1} = \E_\proposal[\weightfunction]$, and $\E_{b_{t+1}}[\testfunction] = \E_\proposal[\weightfunction \testfunction] / \E_\proposal[\weightfunction]$.

The same identity with $b_{\bembed_t}$ in place of $b_t$ gives the Bayes update $\tilde b_{t+1}$ of the generator's belief, with $\E_{\tilde b_{t+1}}[\testfunction] = \E_{\hat\proposal}[\weightfunction\testfunction] / \E_{\hat\proposal}[\weightfunction]$.
The error of the estimate splits into a sampling error and a propagated error,
\begin{equation} \label{eq:error-split}
\tnextestimate - \E_{b_{t+1}}[\testfunction]
= \big(\tnextestimate - \E_{\tilde b_{t+1}}[\testfunction]\big)
+ \big(\E_{\tilde b_{t+1}}[\testfunction] - \E_{b_{t+1}}[\testfunction]\big),
\end{equation}
where the first term comes from drawing only $n$ particles and the second from drawing them from $b_{\bembed_t}$ rather than from $b_t$.

\emph{Propagated error.}
For any bounded continuous $h$ on $\statespace \times \statespace$, let $g(\state) \defeq \int h(\state, \state^\prime)\, \tilde{\transitionfunc}(\state^\prime \mid \state, \obs)\, d\state^\prime$ be the average of $h$ over the successor $\state^\prime$ of a particle at $\state$, so that $\E_{\hat\proposal}[h] = \E_{b_{\bembed_t}}[g]$ and $\E_\proposal[h] = \E_{b_t}[g]$.
The function $g$ is bounded by $\inftynorm{h}$, and it is continuous.
If $\state_k \to \state$, then the successor distributions $\tilde{\transitionfunc}(\cdot \mid \state_k, \obs)$ converge weakly to $\tilde{\transitionfunc}(\cdot \mid \state, \obs)$ by (A3-iii).
The state $\state_k$ is deterministic, so the pair $(\state_k, \state^\prime)$ with $\state^\prime \sim \tilde{\transitionfunc}(\cdot \mid \state_k, \obs)$ converges in distribution to the pair $(\state, \state^\prime)$ with $\state^\prime \sim \tilde{\transitionfunc}(\cdot \mid \state, \obs)$, and $g(\state_k)$, the expectation of $h$ over the former, converges to $g(\state)$.
The inductive hypothesis therefore gives $\E_{\hat\proposal}[h] \to \E_\proposal[h]$ almost surely.
The functions $\weightfunction$ and $\weightfunction\testfunction$ are bounded and continuous (A3-i), hence $\E_{\hat\proposal}[\weightfunction] \to \E_\proposal[\weightfunction] > 0$ (A3-ii) and $\E_{\hat\proposal}[\weightfunction\testfunction] \to \E_\proposal[\weightfunction\testfunction]$ almost surely, and taking the ratio gives $\E_{\tilde b_{t+1}}[\testfunction] \asconv \E_{b_{t+1}}[\testfunction]$.

\emph{Sampling error.}
Lemma~\ref{lem:sllnis}(b) applies with $\proposal_n = \hat\proposal$, since its condition $\liminf_n \E_{\hat\proposal}[\weightfunction] > 0$ holds by the previous paragraph, and gives $\tnextestimate - \E_{\tilde b_{t+1}}[\testfunction] \asconv 0$.
Both terms of Eq.~\ref{eq:error-split} vanish, so $\tnextestimate \asconv \E_{b_{t+1}}[\testfunction]$, which is the convergence of the estimate at step $t+1$.

\emph{Re-embedding.}
The estimate $\tnextestimate$ is the integral of $\testfunction$ against the weighted empirical measure $\hat\mu^{(n)}_{t+1}$ of the new set, which places mass $\particleweight / \sum_j w_j$ on $\particle^\prime$.
Applying the convergence of the estimate to each function in the same class, whose countably many exceptional events have total probability zero, shows that $\hat\mu^{(n)}_{t+1}$ converges weakly to $b_{t+1}$ almost surely.
By (A2), the next embedding depends on the set only through $\hat\mu^{(n)}_{t+1}$, so $b_{\bembed_{t+1}} = \Phi(\hat\mu^{(n)}_{t+1})$, which converges weakly to $\Phi(b_{t+1}) = b_{t+1}$ by weak continuity and well-specification.
This completes the induction.

\paragraph{Almost sure convergence of the sequence.}
Now that we have shown almost sure convergence for any $t \leq \T$, we can complete the proof of the theorem.
For any $\delta > 0$ and $t \leq \T$, there is almost surely a finite random integer $n_t$ such that $|\hat\mu^{(n^\prime)}_t(\testfunction) - \ttargetexpectation| \leq \delta$ for all $n^\prime \geq n_t$.
Let $n^* \defeq \max_{0 \leq t \leq \T} n_t$; then for all $n \geq n^*$ we have $\absdiff \leq \delta$ for every $t \leq \T$.

\end{proof}

\end{theoremrep}

\begin{corollaryrep}[NBF Convergence Rate]
Under the same conditions as Theorem~\ref{thm:nbf-consistency}, suppose additionally that the embedding error is of Monte-Carlo order, $\E_{\generator(\bembed_t)}[g] - \hat\mu^{(n)}_t(g) = O_p(n^{-1/2})$ for every $t \leq \T$ and every bounded continuous $g$, where $\bembed_t$ is the embedding at step $t$ and $\hat\mu^{(n)}_t$ is the weighted empirical measure of the set it embeds.
Then
\[
\maxabsdiff = O_p(n^{-1/2})
\]
as $n \rightarrow \infty$.

\begin{proof}
Use the notation of the proof of Theorem~\ref{thm:nbf-consistency}, and let $R_t$ denote the claim that $\E_{b_{\bembed_t}}[g] - \E_{b_t}[g] = O_p(n^{-1/2})$ for every bounded continuous $g$.
We prove $R_t$ by induction to obtain the rate.

At $t = 0$, $\hat\mu^{(n)}_0(g)$ is an average of $n$ i.i.d.\ terms bounded by $\inftynorm{g}$, so $\hat\mu^{(n)}_0(g) - \E_{\initialstatedist}[g] = O_p(n^{-1/2})$ by Chebyshev's inequality, and adding the hypothesis gives $R_0$.
Assume $R_t$, fix a bounded continuous $\testfunction$, and bound the two terms of Eq.~\ref{eq:error-split}.

\emph{Propagated error.}
Let $g_1(\state) \defeq \int \weightfunction(\state, \state^\prime)\, \testfunction(\state^\prime)\, \tilde{\transitionfunc}(\state^\prime \mid \state, \obs)\, d\state^\prime$ and let $g_2$ be the same integral with $\testfunction$ replaced by one; both are bounded and continuous by the proof of Theorem~\ref{thm:nbf-consistency}.
Then
\[
\E_{\tilde b_{t+1}}[\testfunction] - \E_{b_{t+1}}[\testfunction]
= \frac{\E_{b_{\bembed_t}}[g_1]}{\E_{b_{\bembed_t}}[g_2]} - \frac{\E_{b_t}[g_1]}{\E_{b_t}[g_2]}.
\]
By $R_t$ the two numerators differ by $O_p(n^{-1/2})$, as do the two denominators, and the denominators converge to $\mu_w \defeq \E_\proposal[\weightfunction] > 0$ (A3-ii), so the difference of ratios is $O_p(n^{-1/2})$.

\emph{Sampling error.}
Let $\erri \defeq \particleweight\big(\testfunction(\particle^\prime) - \E_{\tilde b_{t+1}}[\testfunction]\big)$ and $\denominator \defeq \sum_{i=1}^n \particleweight$, so that
\[
\tnextestimate - \E_{\tilde b_{t+1}}[\testfunction] = \frac{\sum_{i=1}^n \erri}{\denominator}.
\]
Conditionally on $b_{\bembed_t}$, the $\erri$ are i.i.d.\ with mean $\E_{\hat\proposal}[\weightfunction\testfunction] - \E_{\hat\proposal}[\weightfunction]\, \E_{\tilde b_{t+1}}[\testfunction] = 0$ and $|\erri| \leq 2 W_{\max} \inftynorm{\testfunction}$, hence
\[
\E\Big[\big(\textstyle\sum_{i=1}^n \erri\big)^2 \,\Big|\, b_{\bembed_t}\Big] \leq 4 n W_{\max}^2 \inftynorm{\testfunction}^2.
\]
Let $\bar\mu_w \defeq \E_{\hat\proposal}[\weightfunction] = \E_{b_{\bembed_t}}[g_2]$, which converges to $\mu_w$ in probability by $R_t$.
The weights lie in $[0, W_{\max}]$, so by Hoeffding's inequality
\[
\Pr\!\left( \tfrac{\denominator}{n} \leq \tfrac{\bar\mu_w}{2} \,\Big|\, b_{\bembed_t} \right) \leq \exp\!\left( -\tfrac{n \bar\mu_w^2}{2 W_{\max}^2} \right).
\]
Let $A_n$ be the event that $\denominator / n > \mu_w/4$.
Its complement requires either $\bar\mu_w \leq \mu_w/2$ or $\denominator/n \leq \bar\mu_w/2$ with $\bar\mu_w > \mu_w/2$, so $\Pr(A_n^c) \leq \Pr(\bar\mu_w \leq \mu_w/2) + \exp(-n\mu_w^2 / 8W_{\max}^2) \to 0$.
On $A_n$ the sampling error is at most $\tfrac{4}{\mu_w n} \big|\sum_i \erri\big|$ in absolute value, so Chebyshev's inequality gives
\[
\Pr\!\left[\sqrt{n}\, \big|\tnextestimate - \E_{\tilde b_{t+1}}[\testfunction]\big| \geq M\right] \leq \Pr(A_n^c) + \frac{64\, W_{\max}^2 \inftynorm{\testfunction}^2}{\mu_w^2 M^2},
\]
so the sampling error is $O_p(n^{-1/2})$.

Both terms of Eq.~\ref{eq:error-split} are $O_p(n^{-1/2})$, so $\tnextestimate - \E_{b_{t+1}}[\testfunction] = O_p(n^{-1/2})$, which is the rate at step $t+1$.
Applying it with $g$ in place of $\testfunction$ and adding the hypothesis at step $t+1$,
\[
\E_{b_{\bembed_{t+1}}}[g] - \E_{b_{t+1}}[g]
= \big(\E_{b_{\bembed_{t+1}}}[g] - \hat\mu^{(n)}_{t+1}(g)\big) + \big(\hat\mu^{(n)}_{t+1}(g) - \E_{b_{t+1}}[g]\big)
= O_p(n^{-1/2}) + O_p(n^{-1/2}),
\]
which is $R_{t+1}$.
A maximum of finitely many $O_p(n^{-1/2})$ terms is $O_p(n^{-1/2})$, which gives the supremum over $t \leq \T$.
\end{proof}
\end{corollaryrep}

\section{Domains}
\label{app:domains}
\subsection{Pursuit--evasion}
\label{app:pe-domain}
The arena is the square $[0, L]^2$ with $L = 10$ and a single stationary range-only sensor at $\mathbf{x}_{\text{sensor}} = (5, 5)$.
Three rectangular obstacles create a rooms-and-corridors layout: a left vertical wall $[2, 3]\times[0, 7]$ with a gap above it, a right vertical wall $[7, 8]\times[3, 10]$ with a gap below it, and a central block $[4, 6]\times[4, 6]$.

At each step every evader's policy draws its action, a heading $\action$; the next position is $x' = x + v(\cos\action, \sin\action) + \epsilon$ with $\epsilon \sim \mc N(0, \sigma_{\text{dyn}}^2 I)$, and the move is integrated through $N_{\text{sub}}$ substeps with reflective wall and obstacle clipping.
For $k > 1$, evaders are pushed away from each other according to the repulsion strength $s_\text{int}$ if they are within each other's radius $r_\text{int}$.

Each evader produces a per-step range observation $y_j = \|x_j - \mathbf{x}_{\text{sensor}}\| + \mc N(0, \sigma^2)$ with $\sigma$ drawn once per episode from $\mc U[\sigma_{\min}, \sigma_{\max}]$.
The per-evader observations are randomly permuted before being passed to the filter, so the index of an observation carries no information about which evader produced it.
This forces the filter to maintain a posterior over the joint state rather than $k$ independent marginals.
Only the evaders' policies, initial positions and headings, and the observation noise $\sigma$ vary between episodes.
All other quantities, including obstacle positions, are fixed across training and evaluation.
The exact settings are summarized in Table~\ref{tab:pe-env}.

\begin{table}[ht]
  \centering
  \small
  \caption{Pursuit--evasion environment settings.}
  \label{tab:pe-env}
  \begin{tabular}{@{}lll@{}}
    \toprule
    Symbol & Value & Description \\
    \midrule
    $L$                     & $10.0$            & Grid size.\\
    $k$                     & $\{1, 3, 5\}$  & Number of evaders.\\
    $\T$                    & $20$              & Episode length (steps). \\
    $v$                     & $0.5$             & Per-step speed. \\
    $\sigma_{\text{dyn}}$   & $0.2$             & Dynamics noise std (isotropic). \\
    $N_{\text{sub}}$        & $4$               & Substeps per move, each clipped against walls and obstacles. \\
    $\mathbf{x}_{\text{sensor}}$ & $(5, 5)$     & Sensor position. \\
    $\sigma_{\min}, \sigma_{\max}$ & $0.01, 0.05$ & Range-noise std bounds. \\
    $r_{\text{int}}$        & $1.5$             & Pairwise repulsion radius between evaders. \\
    $s_{\text{int}}$        & $1.0$             & Pairwise repulsion strength. \\
    \bottomrule
  \end{tabular}
\end{table}

\subsubsection{Policies}
\label{app:policies}

Every evader is controlled by a stochastic heading policy built from one of five primitives.
Table~\ref{tab:policy-kinds} gives the heading rule of each primitive and Table~\ref{tab:policy-mixes} the training mixture and the held-out policies.

\begin{table}[ht]
    \centering
    \small
    \caption{Policy primitives; the action $\action$ is the heading and $\eta \sim \mc N(0, 1)$ a fresh draw at every step.}
    \label{tab:policy-kinds}
    \begin{tabular}{@{}ll@{}}
        \toprule
        Primitive & Heading rule \\
        \midrule
        \texttt{goal}$(\mathbf{g}, c)$ & $\action = \operatorname{atan2}(\mathbf{g} - x) + \eta / c$ \\
        \texttt{random\_walk}          & $\action \sim \mc U[0, 2\pi)$ \\
        \texttt{patrol}$(\{\mathbf{p}_i\}, c)$ & $\action = \operatorname{atan2}(\mathbf{p} - x) + \eta / c$ toward the nearest waypoint $\mathbf{p}$ farther than $0.5$ \\
        \texttt{evasive}               & $\action = \operatorname{atan2}(x - \mathbf{x}_{\text{sensor}}) + 0.3\, \eta$ \\
        \texttt{wall\_hugger}          & $\action$ along the nearest wall $+\, 0.2\, \eta$ \\
        \bottomrule
    \end{tabular}
\end{table}

\begin{table}[ht]
    \centering
    \small
    \caption{Training mixture and held-out policies.}
    \label{tab:policy-mixes}
    \begin{tabular}{@{}lcl@{}}
        \toprule
        Policy & Weight & Parameters \\
        \midrule
        \multicolumn{3}{l}{\emph{Training mixture}} \\
        \texttt{goal}        & 40\% & $\mathbf{g} \sim \mc U[0, L]^2$, $c \sim \mc U(1, 5)$ \\
        \texttt{random\_walk}& 20\% & --- \\
        \texttt{patrol}      & 20\% & $n_{\text{wp}} \sim \mc U\{2,3,4\}$, $\mathbf{p}_i \sim \mc U[0.5, L\!-\!0.5]^2$, $c = 3$ \\
        \texttt{evasive}     & 20\% & --- \\
        \midrule
        \multicolumn{3}{l}{\emph{Held-out policies}} \\
        \texttt{opposite\_corner\_A} & --- & \texttt{goal}$((1, 1), 5)$ \\
        \texttt{opposite\_corner\_B} & --- & \texttt{goal}$((9, 9), 5)$ \\
        \texttt{patrol\_diamond}     & --- & \texttt{patrol}$(\{(2, 5), (5, 8), (8, 5), (5, 2)\}, 3)$ \\
        \texttt{wall\_hugger}        & --- & --- \\
        \bottomrule
    \end{tabular}
\end{table}

Training draws each evader's policy independently from the mixture.
The in-distribution evaluation runs one training-mixture component per cell, drawn with the cell's seed and shared by the $k$ evaders, so that each policy's columns in Table~\ref{tab:pe-policies} isolate one belief family.
Table~\ref{tab:pe-main} and the rest of the experiments in Appendix~\ref{app:add-experiments} average the four with equal weight.
The held-out policies are evaluated at $k = 1$ and $k = 5$.

\subsection{Lorenz-96}
\label{app:l96-domain}
The state is $\state \in \R^K$ with dynamics $\dot{x}_j = (x_{j+1} - x_{j-2})\, x_{j-1} - x_j + F$, where the index $j$ is taken modulo $K$ so that the coordinates form a ring.
Each window integrates the dynamics with $N_{\text{sub}}$ RK4 steps of size $\Delta t$ and adds Gaussian noise $\epsilon \sim \mc N(0, \sigma_{\text{dyn}}^2 I)$, so the transition kernel is stochastic.
The odd-indexed sites are observed, $y_j = x_j + \mc N(0, \sigma_{\text{obs}}^2)$ for $j \in \{1, 3, \dots, K - 1\}$.
During training $\sigma_{\text{obs}}$ is drawn once per episode from $\mc U[\sigma_{\min}, \sigma_{\max}]$; at evaluation it is $1.0$, and the likelihood sweep of Section~\ref{sec:experiments} sets it to each value in $\{0.25, 0.5, 1.0, 1.5, 2.0\}$.
An episode draws $F$ from one of the two training bands (Table~\ref{tab:l96-env}) and starts from $F + \mc N(0, 0.25\, I)$.
The trajectory is integrated for $t_{\text{spin}}$ time units before the first window so that it has settled onto the attractor, the region of the state space that the chaotic dynamics visit in the long run.
The initial belief is $\mc N(\statetime{0} + \delta, \sigma_{\text{init}}^2 I)$ with $\delta \sim \mc N(0, I)$ drawn at the start of the episode.
The exact settings are found in Table~\ref{tab:l96-env}.
The window of 0.4 time units is eight times the 0.05 step of the standard benchmark \citep{lorenz1996predictability}.
At 0.05 the forecast is close to linear and a localized ensemble Kalman filter is near-optimal, while at 0.4 the chaotic dynamics run long enough between observations that the forecast ensemble is non-Gaussian.

\begin{table}[ht]
  \centering
  \small
  \caption{Lorenz-96 environment settings.}
  \label{tab:l96-env}
  \begin{tabular}{@{}lll@{}}
    \toprule
    Symbol & Value & Description \\
    \midrule
    $K$                          & $\{20, 40\}$      & State dimension. \\
    $\T$                         & $20$              & Episode length (windows). \\
    $\Delta t$, $N_{\text{sub}}$ & $0.01$, $40$      & RK4 step and steps per window (window $0.4$). \\
    $\sigma_{\text{dyn}}$        & $0.25$            & Process noise std per window. \\
    $\sigma_{\min}, \sigma_{\max}$ & $0.5, 1.1$      & Observation-noise std bounds (training). \\
    $F$ (training)               & $[5, 7] \cup [9, 11]$ & Forcing bands. \\
    $F$ (evaluation)             & $\{6, 8, 10\}$    & In-range and held-out forcings. \\
    $t_{\text{spin}}$            & $10$              & Spin-up (time units). \\
    $\sigma_{\text{init}}$       & $2.0$             & Initial belief std. \\
    \bottomrule
  \end{tabular}
\end{table}

The particle weights in this domain are the vectors of standardized residuals of Section~\ref{sec:experiments}.
The scalar weight of Algorithm~\ref{alg:neural-filtering}, for which Theorem~\ref{thm:nbf-consistency} is stated, is a function of the residual vector, so by Proposition~\ref{prop:realized-state}, the set posterior of the vector-weighted set scores at least as well in expectation as that of the scalar-weighted set of the same particles.

The residual is standardized by $\sigma_{\text{obs}}$, which is not given to the encoder, but the set still determines the observation.
Two particles $i$ and $k$ that differ at an observed site $j$ give $\sigma_{\text{obs}} = (\state^\prime_{k,j} - \state^\prime_{i,j}) / (r_{i,j} - r_{k,j})$ from their residuals $r$, and then $\obs_j = \state^\prime_{i,j} + \sigma_{\text{obs}}\, r_{i,j}$, so Lemma~\ref{lem:set-determines-obs} applies at every noise level.
Covariance inflation and localization \citep{gaspari1999construction,hunt2007efficient} scale the ensemble Kalman update to this dimension, and local analyses \citep{poterjoy2016localized,schenk2022two} scale the particle filter, in both cases by restricting each update to nearby sites.
Our implementation of EnKF is localized and inflated in the same way, and NBF's encoder is passed the whole state and residual vectors with no localization imposed.

\section{Architecture and Training Details}
\label{app:training}
The same model is trained on both domains: a Set Transformer encoder that maps a weighted particle set to $\bembed$ and a conditional flow-matching generator.
One model is trained per state dimension ($K$ on Lorenz-96, $k$ on pursuit--evasion), with the sizes in Table~\ref{tab:nbf-hparams}.
Hyperparameters were not tuned extensively.

\paragraph{Encoder.}
The encoder input for each particle is its state concatenated with its weight feature: on pursuit--evasion the scalar weight of Algorithm~\ref{alg:neural-filtering} as a max-shifted log-likelihood, floored at $-30$ nats, and on Lorenz-96 the $K$-vector of standardized residuals.
The weight enters only as this feature, with no resampling or weighted pooling during propagation or embedding.
The set is processed by three rounds of an Induced Set Attention Block \citep{lee2019set}, each of which compresses the $n$ particles into $M$ learned inducing points by cross-attention and broadcasts the summary back, at $O(nM)$ cost per block, followed by a pool over the particles whose output is concatenated to every particle's input for the next round.
The pool is a mean over the particles on pursuit--evasion and a single-query attention pool \citep{lee2019set} on Lorenz-96, and the final pool produces $\bembed \in \R^{d_{\text{emb}}}$.
States are standardized before they enter the encoder and the generator, on Lorenz-96 by the climatological mean and standard deviation of the state estimated from 512 simulated episodes and on pursuit--evasion by the affine map of $[0, L]^2$ onto $[-1, 1]^2$. Every reported NLL is in the unnormalized state units.

\paragraph{Generative model.}
The generator is a conditional flow-matching model whose velocity network is an MLP on the concatenation of the state, a time embedding, and $\bembed$.
Sampling integrates the flow from $z \sim \mc N(0, I)$ with a fixed-step midpoint method, and the density integrates the flow and its exact divergence backward with Euler steps.

\paragraph{Training.}
Training follows Section~\ref{sec:training}.
Every cycle simulates 64 episodes of $\T = 20$ steps, runs NBF along them with the current parameters and $n$ particles, and takes 40 gradient steps on batches of 256 (set, realized state) pairs drawn from those rollouts.
The optimizer is AdamW with a linear warmup of 5{,}000 steps and cosine decay from $10^{-4}$ to $10^{-6}$, with gradients clipped at norm $1.0$; gradients flow through the encoder and the generator and not through the rollout.
During rollouts the generator samples with 10 midpoint steps.
The same 10-step generator is the one the deployed filter uses at evaluation; 100 Euler steps are used to score NLL.
Training runs for 100{,}000 total steps.

\paragraph{Lorenz-96.}
On Lorenz-96 a particle's weight is the vector of standardized residuals at the observed sites (Section~\ref{sec:experiments}), clipped to $[-20, 20]$ and zero at the unobserved sites.
The rest of the model is unchanged.
Each training episode draws $F$ from the training bands and $\sigma_{\text{obs}}$ from $[0.5, 1.1]$ (Table~\ref{tab:l96-env}).
Rollouts use $n = 4096$ particles at $K = 40$, with each gradient step accumulated over 8 micro-batches, and $n = 2048$ at $K = 20$, with the velocity network in bfloat16 during rollouts.
Three different seeds were used to train models, each on one NVIDIA L40S GPU.
One model trains in 16 hours at $K = 20$ and in 36 hours at $K = 40$, the same to within 0.1 hours across the three seeds.

\paragraph{Pursuit--evasion.}
Rollouts use $n = 2048$ particles at every $k$ with the velocity network in bfloat16, and the deployed filter is the 10-step bfloat16 generator as on Lorenz-96.
Each training episode draws $k$ policies from the training mixture (Table~\ref{tab:policy-mixes}) and $\sigma$ from $[0.01, 0.05]$.
Three different seeds were used to train models, each on one NVIDIA L40S GPU.
One model trains in 7.7 hours at $k = 1$, 14.4 hours at $k = 3$ and 16.4 hours at $k = 5$, the same to within 0.2 hours across the three seeds.

\begin{table}[ht]
  \centering
  \small
  \caption{NBF encoder, generator, and training hyperparameters per model.}
  \label{tab:nbf-hparams}
  \begin{tabular}{@{}lccccc@{}}
    \toprule
    & \multicolumn{2}{c}{Lorenz-96} & \multicolumn{3}{c}{Pursuit--evasion} \\
    \cmidrule(lr){2-3} \cmidrule(lr){4-6}
    & $K=20$ & $K=40$ & $k=1$ & $k=3$ & $k=5$ \\
    \midrule
    \multicolumn{6}{l}{\emph{Encoder (Set Transformer)}} \\
    Encoder input width            & 40  & 80  & 3   & 7   & 11 \\
    Embedding dim $d_{\text{emb}}$ & 192 & 256 & 64  & 128 & 128 \\
    Hidden size                    & 512 & 512 & 256 & 256 & 256 \\
    Rounds                         & \multicolumn{5}{c}{3} \\
    Heads                          & \multicolumn{5}{c}{8} \\
    Inducing points $M$            & 64  & 64  & 32  & 64  & 64 \\
    Pool                           & attention & attention & mean & mean & mean \\
    \midrule
    \multicolumn{6}{l}{\emph{Generative model (flow matching)}} \\
    MLP hidden size                & 512 & 512 & 256 & 512 & 512 \\
    MLP layers                     & 8   & 8   & 6   & 6   & 8 \\
    Time-embedding dim             & \multicolumn{5}{c}{16} \\
    Sampling steps (rollout / deployed) & \multicolumn{5}{c}{10 / 10} \\
    Density steps                  & \multicolumn{5}{c}{100} \\
    \midrule
    \multicolumn{6}{l}{\emph{Training}} \\
    Rollout particles $n$          & 2048 & 4096 & 2048 & 2048 & 2048 \\
    Episodes per cycle             & \multicolumn{5}{c}{64} \\
    Gradient steps per cycle       & \multicolumn{5}{c}{40} \\
    Batch size                     & \multicolumn{5}{c}{256} \\
    Gradient steps                 & \multicolumn{5}{c}{100k} \\
    Peak / final LR                & \multicolumn{5}{c}{$10^{-4}$ / $10^{-6}$, 5000 warmup steps, cosine decay} \\
    Grad-clip norm                 & \multicolumn{5}{c}{1.0} \\
    \bottomrule
  \end{tabular}
\end{table}

\section{Baselines}
\label{app:baselines}
\paragraph{Particle filters.}
PF is the bootstrap filter \citep{gordon1993novel}: each particle is simulated through the transition, reweighted by the observation likelihood, and the set is resampled systematically whenever the effective sample size $1/\sum_i w_i^2$ of the normalized weights falls below $n/2$.
NBF runs with the bootstrap proposal in every experiment; the variants below change the proposal or add a move after resampling, and any of them could be used as a drop-in replacement in Algorithm~\ref{alg:neural-filtering}.
PF-Reg, PF-RM and PF-OP share the initialization and resampling of PF.
PF-Reg jitters the resampled set with the shrinkage kernel of \citet{liu2001combined}: particles are contracted toward the weighted mean by $\sqrt{1 - h^2}$ and perturbed with covariance $h^2$ times the pre-resample weighted covariance, so the ensemble covariance is preserved.
The bandwidth $h$ is chosen according to Silverman's rule \citep{silverman1986density} for the particle count and state dimension.
PF-RM, the resample-move filter of \citet{gilks2001following}, follows each resampling with an MCMC move that leaves the posterior invariant.
On Lorenz-96 the move is an exact draw from $p(\state^\prime \mid \state, \obs)$ given the particle's ancestor $\state$, which is Gaussian because the transition adds Gaussian noise to the integrated state and the observation is Gaussian on the observed sites.
PF-OP draws every particle from the same conditional and weights it by the marginal likelihood $p(\obs \mid \state)$, the locally optimal proposal of \citet{doucet2009tutorial}.

\paragraph{Ensemble Kalman filters.}
EnKF is the stochastic ensemble Kalman filter with perturbed observations \citep{evensen1994sequential,burgers1998analysis}, with $m$ equal-weight members, multiplicative inflation of $1.05$ on the member deviations before the analysis, and the Gaspari--Cohn taper \citep{gaspari1999construction} on the sample covariance with a localization radius of 4 sites.
The filter was validated against published Lorenz results: in the fully observed $K = 40$, window $0.05$ configuration of the DAPPER suite \citep{raanes2024dapper} it reaches an analysis RMSE of $0.200$ at 40 members against the reported $0.22$, and in the half-observed, window $0.3$ configuration of \citet{schenk2022two} it reaches $0.840$ between their LMCPF at $0.77$ and their LETKF at $0.86$.
Its NLL is the log-density of the true state under the Gaussian with the ensemble mean and the tapered ensemble covariance.
On pursuit--evasion the observation is unlabelled, so each member's predicted ranges and the shuffled observation are sorted before the update, and no localization is applied.

\paragraph{Gaussian sum filter.}
GMF carries $m$ weighted Gaussian components \citep{alspach1972nonlinear}, each updated as an unscented Kalman filter \citep{julier2004unscented} whose forecast propagates the sigma points with the policy's mean heading and whose analysis matches the sorted observed ranges in rank order to the ranges predicted from the component mean.
The mixture weights are multiplied by each component's likelihood of the observation and renormalized.
After every update, pairs of components whose symmetric KL divergence is below $0.5$ are merged by moment matching and components below a weight of $0.01$ are dropped, so the count stays at most $m$.
The components start at the arena centre with jitter of scale $L/10$ and covariance $(L/2)^2 I$, and the NLL is the log-density of the true state under the mixture.

\section{Additional Experiments}
\label{app:add-experiments}

\paragraph{Mechanism.}
Within one step, NBF's encoder receives the same propagated, weighted particles that a particle filter would resample.
Table~\ref{tab:pe-mechanism} scores NBF's propagated set at every step in two ways, as a \emph{weighted draw} under its importance weights, which is what a particle filter reports, and after \emph{embed-and-regenerate}, which is what NBF reports.
At $k = 1$ the two reads score alike, since the weights still carry the posterior.
At $k = 5$ the effective sample size is about two particles, so the weighted draw is close to a point mass, while embed-and-regenerate of the same particles gives a calibrated belief.

\begin{table}[ht]
    \centering
    \footnotesize
    \setlength{\tabcolsep}{3pt}
    \caption{Pursuit--evasion, NBF's propagated set with $n = 2048$ and $\sigma = 0.02$, scored as a weighted draw and after embed-and-regenerate. ESS/$n$ is the effective sample size as a fraction of $n$.}
    \label{tab:pe-mechanism}
    \begin{tabular}{lrrrr}
\toprule
Read of the set & CRPS & cov50 & cov90 & ESS/$n$ \\
\midrule
\multicolumn{5}{l}{$k = 1$} \\
weighted draw & 1.266 $\pm$ 0.020 & 0.47 & 0.87 & 0.129 \\
embed-and-regenerate & 1.284 $\pm$ 0.020 & 0.55 & 0.93 & 0.129 \\
\midrule
\multicolumn{5}{l}{$k = 5$} \\
weighted draw & 1.833 $\pm$ 0.012 & 0.02 & 0.04 & 0.001 \\
embed-and-regenerate & 0.924 $\pm$ 0.007 & 0.60 & 0.97 & 0.001 \\
\bottomrule
\end{tabular}

\end{table}

\paragraph{Calibration metrics.}
CRPS and coverage are computed from 1024 equal-weight samples of every filter at every step, drawn from the generator of the updated embedding for NBF and by weight from the particles or members of the classical filters.
Coverage (cov50, cov90) is the fraction of true coordinates inside the central 50\% and 90\% intervals of the filter's marginals, 0.50 and 0.90 for a calibrated filter and lower for an overconfident one.
On pursuit--evasion with $k > 1$, the shuffled observations make the belief symmetric under relabelling of the evaders.
Before CRPS and coverage are computed, every method's samples are relabelled by the assignment of evaders to the true ones that minimizes the squared distance.

\begin{figure}[ht]
    \centering
    \includegraphics[width=\linewidth]{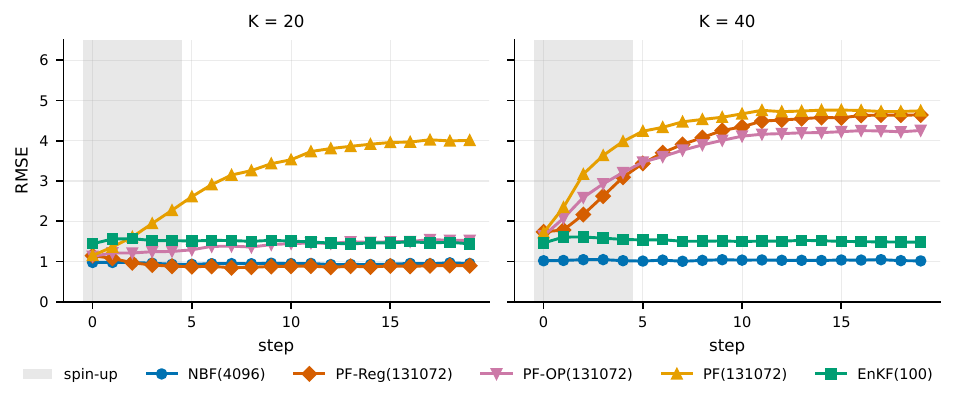}
    \caption{RMSE against the step of the episode at the held-out $F = 8$, averaged over episodes. The shaded area indicates spin-up time.}
    \label{fig:l96-per-step}
\end{figure}

\paragraph{Per-step error.}
Figure~\ref{fig:l96-per-step} plots RMSE against the step of the episode at the held-out $F = 8$.

\begin{figure}[ht]
    \centering
    \includegraphics[width=\linewidth]{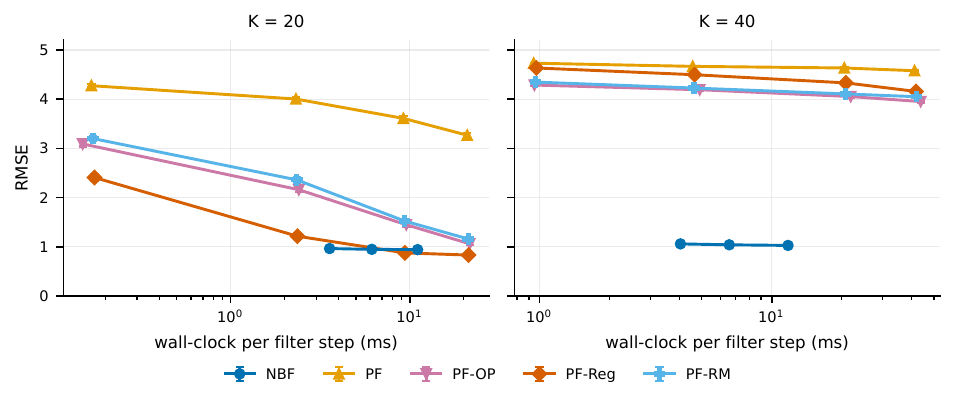}
    \caption{RMSE at the held-out $F = 8$ against wall-clock per filter step; each line joins the budgets of one family, from 1024 to 4096 particles for NBF and from 8192 to 262144 particles for the particle filters.}
    \label{fig:l96-compute}
\end{figure}

\paragraph{Compute-matched comparison.}
Figure~\ref{fig:l96-compute} plots RMSE against wall-clock per filter step for NBF and every particle-filter family.

\paragraph{Held-out settings.}
Tables~\ref{tab:l96-forcing} and~\ref{tab:l96-sigma} report every method on Lorenz-96 across the forcing and the observation noise, and Table~\ref{tab:pe-ood} reports the main methods on the individual held-out pursuit--evasion policies.

\begin{table}[ht]
    \centering
    \scriptsize
    \setlength{\tabcolsep}{3pt}
    \caption{Lorenz-96 RMSE across the forcing $F$ at $\sigma_{\text{obs}} = 1$; training draws $F$ from $[5, 7] \cup [9, 11]$. SEMs are at most 0.09.}
    \label{tab:l96-forcing}
    \begin{tabular}{lrrrrrr}
\toprule
 & \multicolumn{3}{c}{$K = 20$} & \multicolumn{3}{c}{$K = 40$} \\
\cmidrule(lr){2-4} \cmidrule(lr){5-7}
Method & $F = 6$ & $F = 8$ & $F = 10$ & $F = 6$ & $F = 8$ & $F = 10$ \\
\midrule
NBF(1024) & 0.769 & 0.965 & 1.162 & 0.824 & 1.060 & 1.365 \\
NBF(2048) & 0.762 & 0.949 & 1.125 & 0.818 & 1.041 & 1.311 \\
NBF(4096) & 0.760 & 0.943 & 1.107 & 0.814 & 1.029 & 1.273 \\
EnKF(50) & 0.939 & 1.503 & 2.192 & 0.960 & 1.524 & 2.274 \\
EnKF(100) & 0.933 & 1.484 & 2.166 & 0.953 & 1.506 & 2.234 \\
PF(8192) & 1.971 & 4.275 & 5.640 & 3.045 & 4.732 & 5.899 \\
PF(32768) & 1.476 & 4.004 & 5.488 & 2.905 & 4.668 & 5.871 \\
PF(131072) & 1.093 & 3.611 & 5.324 & 2.735 & 4.635 & 5.839 \\
PF(262144) & 0.939 & 3.268 & 5.228 & 2.639 & 4.579 & 5.806 \\
PF-OP(8192) & 0.965 & 3.091 & 4.942 & 2.347 & 4.287 & 5.543 \\
PF-OP(32768) & 0.777 & 2.161 & 4.467 & 2.078 & 4.192 & 5.464 \\
PF-OP(131072) & 0.701 & 1.449 & 3.394 & 1.813 & 4.054 & 5.407 \\
PF-OP(262144) & 0.694 & 1.062 & 2.716 & 1.597 & 3.952 & 5.375 \\
PF-Reg(8192) & 0.799 & 2.408 & 4.769 & 2.638 & 4.635 & 5.865 \\
PF-Reg(32768) & 0.703 & 1.216 & 3.139 & 2.201 & 4.497 & 5.817 \\
PF-Reg(131072) & 0.684 & 0.877 & 1.553 & 1.558 & 4.332 & 5.773 \\
PF-Reg(262144) & 0.682 & 0.832 & 1.132 & 1.247 & 4.158 & 5.712 \\
PF-RM(8192) & 1.063 & 3.201 & 5.015 & 2.465 & 4.348 & 5.572 \\
PF-RM(32768) & 0.801 & 2.361 & 4.551 & 2.188 & 4.227 & 5.516 \\
PF-RM(131072) & 0.716 & 1.526 & 3.618 & 1.884 & 4.108 & 5.466 \\
PF-RM(262144) & 0.702 & 1.160 & 2.983 & 1.765 & 4.051 & 5.434 \\
\bottomrule
\end{tabular}

\end{table}

\begin{table}[ht]
    \centering
    \scriptsize
    \setlength{\tabcolsep}{3pt}
    \caption{Lorenz-96 RMSE and NLL at the held-out $F = 8$ across the observation noise $\sigma_{\text{obs}}$. Training draws $\sigma_{\text{obs}}$ from $[0.5, 1.1]$. SEMs are at most 0.09 on RMSE and 2.1 nats on NLL.}
    \label{tab:l96-sigma}
    \begin{tabular}{lrrrrrrrrrr}
\toprule
 & \multicolumn{5}{c}{$K = 20$} & \multicolumn{5}{c}{$K = 40$} \\
\cmidrule(lr){2-6} \cmidrule(lr){7-11}
Method & $0.25$ & $0.5$ & $1$ & $1.5$ & $2$ & $0.25$ & $0.5$ & $1$ & $1.5$ & $2$ \\
\midrule
\multicolumn{11}{l}{RMSE} \\
NBF(1024) & 0.521 & 0.561 & 0.965 & 1.454 & 2.031 & 2.337 & 0.617 & 1.060 & 1.569 & 2.117 \\
NBF(2048) & 0.463 & 0.551 & 0.949 & 1.434 & 2.008 & 2.070 & 0.600 & 1.041 & 1.542 & 2.090 \\
NBF(4096) & 0.433 & 0.547 & 0.943 & 1.420 & 1.997 & 1.770 & 0.592 & 1.029 & 1.525 & 2.069 \\
EnKF(50) & 0.570 & 0.903 & 1.503 & 1.918 & 2.214 & 0.591 & 0.926 & 1.524 & 1.940 & 2.242 \\
EnKF(100) & 0.557 & 0.886 & 1.484 & 1.898 & 2.197 & 0.570 & 0.907 & 1.506 & 1.922 & 2.220 \\
PF(8192) & 4.686 & 4.547 & 4.275 & 3.940 & 3.706 & 4.855 & 4.841 & 4.732 & 4.595 & 4.417 \\
PF(32768) & 4.634 & 4.453 & 3.993 & 3.583 & 3.304 & 4.832 & 4.797 & 4.667 & 4.498 & 4.320 \\
PF(131072) & 4.564 & 4.294 & 3.622 & 3.119 & 2.856 & 4.799 & 4.764 & 4.635 & 4.413 & 4.172 \\
PF(262144) & 4.522 & 4.140 & 3.274 & 2.772 & 2.612 & 4.805 & 4.752 & 4.582 & 4.331 & 4.118 \\
PF-OP(8192) & 1.495 & 2.716 & 3.091 & 3.086 & 2.972 & 2.636 & 3.776 & 4.287 & 4.260 & 4.143 \\
PF-OP(32768) & 1.085 & 1.814 & 2.159 & 2.361 & 2.391 & 2.552 & 3.721 & 4.190 & 4.119 & 3.929 \\
PF-OP(131072) & 0.721 & 1.280 & 1.456 & 1.688 & 1.960 & 2.409 & 3.648 & 4.057 & 3.930 & 3.744 \\
PF-OP(262144) & 0.604 & 0.854 & 1.075 & 1.443 & 1.759 & 2.333 & 3.578 & 3.964 & 3.813 & 3.642 \\
PF-Reg(8192) & 4.683 & 4.367 & 2.423 & 1.663 & 1.696 & 4.869 & 4.819 & 4.635 & 4.260 & 3.704 \\
PF-Reg(32768) & 4.597 & 3.815 & 1.234 & 1.239 & 1.550 & 4.844 & 4.761 & 4.494 & 3.850 & 3.095 \\
PF-Reg(131072) & 4.509 & 2.717 & 0.900 & 1.154 & 1.508 & 4.824 & 4.741 & 4.333 & 3.185 & 2.369 \\
PF-Reg(262144) & 4.390 & 1.922 & 0.836 & 1.136 & 1.499 & 4.774 & 4.714 & 4.131 & 2.706 & 2.090 \\
PF-RM(8192) & 1.877 & 2.957 & 3.201 & 3.102 & 3.016 & 2.714 & 3.833 & 4.348 & 4.300 & 4.136 \\
PF-RM(32768) & 1.653 & 2.452 & 2.351 & 2.389 & 2.464 & 2.666 & 3.797 & 4.230 & 4.156 & 3.971 \\
PF-RM(131072) & 1.298 & 1.639 & 1.506 & 1.729 & 1.988 & 2.576 & 3.731 & 4.108 & 3.950 & 3.753 \\
PF-RM(262144) & 1.115 & 1.216 & 1.154 & 1.428 & 1.758 & 2.545 & 3.678 & 4.051 & 3.851 & 3.624 \\
\midrule
\multicolumn{11}{l}{NLL} \\
NBF(1024) & 11.9 & 13.2 & 23.4 & 32.1 & 41.7 & 113.6 & 32.8 & 53.6 & 71.3 & 90.6 \\
NBF(2048) & 9.8 & 12.9 & 23.2 & 31.8 & 41.3 & 103.0 & 32.1 & 52.9 & 70.5 & 89.6 \\
NBF(4096) & 8.8 & 12.8 & 23.1 & 31.6 & 41.1 & 90.4 & 31.7 & 52.6 & 70.0 & 88.8 \\
EnKF(50) & 10.8 & 20.6 & 32.1 & 38.3 & 41.9 & 20.8 & 40.9 & 64.0 & 76.3 & 83.9 \\
EnKF(100) & 9.5 & 19.8 & 31.7 & 37.8 & 41.6 & 18.1 & 39.6 & 63.1 & 75.5 & 83.2 \\
\bottomrule
\end{tabular}

\end{table}

\begin{table}[ht]
    \centering
    \footnotesize
    \setlength{\tabcolsep}{3pt}
    \caption{Pursuit--evasion held-out policies at $k = 1$ and $k = 5$ for the methods of Table~\ref{tab:pe-main}. SEMs are at most 0.06 on CRPS.}
    \label{tab:pe-ood}
    \begin{tabular}{lrrrr}
\toprule
 & \multicolumn{2}{c}{$k = 1$} & \multicolumn{2}{c}{$k = 5$} \\
\cmidrule(lr){2-3} \cmidrule(lr){4-5}
Method & NLL & CRPS & NLL & CRPS \\
\midrule
\multicolumn{5}{l}{\emph{corner A}} \\
NBF(512) & -0.39 & 0.424 & 20.8 & 0.947 \\
NBF(2048) & -0.45 & 0.404 & 19.9 & 0.959 \\
PF(32768) & -- & 0.208 & -- & 1.076 \\
PF-Reg(32768) & -- & 0.257 & -- & 1.133 \\
EnKF(100) & 3.15 & 0.710 & 21.3 & 0.513 \\
GMF(64) & 10.9 & 0.818 & 333.2 & 0.668 \\
\midrule
\multicolumn{5}{l}{\emph{corner B}} \\
NBF(512) & -0.46 & 0.403 & 20.3 & 0.891 \\
NBF(2048) & -0.51 & 0.390 & 19.6 & 0.893 \\
PF(32768) & -- & 0.204 & -- & 1.131 \\
PF-Reg(32768) & -- & 0.249 & -- & 1.147 \\
EnKF(100) & 3.25 & 0.680 & 21.2 & 0.526 \\
GMF(64) & 9.32 & 0.785 & 351.3 & 0.711 \\
\midrule
\multicolumn{5}{l}{\emph{patrol diamond}} \\
NBF(512) & 0.51 & 1.022 & 23.4 & 0.734 \\
NBF(2048) & 0.48 & 1.026 & 20.8 & 0.719 \\
PF(32768) & -- & 0.898 & -- & 1.258 \\
PF-Reg(32768) & -- & 0.925 & -- & 1.298 \\
EnKF(100) & 4.44 & 1.187 & 24.5 & 0.729 \\
GMF(64) & 4.62 & 1.275 & 357.0 & 0.994 \\
\midrule
\multicolumn{5}{l}{\emph{wall hugger}} \\
NBF(512) & 0.51 & 2.119 & 22.1 & 1.280 \\
NBF(2048) & 0.39 & 2.085 & 20.9 & 1.278 \\
PF(32768) & -- & 0.662 & -- & 2.150 \\
PF-Reg(32768) & -- & 1.070 & -- & 2.147 \\
EnKF(100) & 6.63 & 1.888 & 40.1 & 1.155 \\
GMF(64) & 6.12 & 2.414 & 2161.7 & 1.486 \\
\bottomrule
\end{tabular}

\end{table}

\paragraph{Expanded results.}
Table~\ref{tab:l96-ladder} reports every method and budget on Lorenz-96 at the held-out $F = 8$, including the resample-move and locally optimal proposal filters.
Table~\ref{tab:pe-ladder} reports every method on pursuit--evasion, averaged over the four training-mixture policies as in Table~\ref{tab:pe-main}.
Table~\ref{tab:pe-policies} reports each policy separately.

\begin{table}[ht]
    \centering
    \scriptsize
    \setlength{\tabcolsep}{3pt}
    \caption{Lorenz-96 at the held-out $F = 8$, every method and budget; ms is wall-clock per filter step. SEMs are at most 0.3 nats on NLL and 0.07 on RMSE and CRPS.}
    \label{tab:l96-ladder}
    \begin{tabular}{lrrrrrrrrrrrr}
\toprule
 & \multicolumn{6}{c}{$K = 20$} & \multicolumn{6}{c}{$K = 40$} \\
\cmidrule(lr){2-7} \cmidrule(lr){8-13}
Method & NLL & RMSE & CRPS & cov50 & cov90 & ms & NLL & RMSE & CRPS & cov50 & cov90 & ms \\
\midrule
NBF(1024) & 23.4 & 0.965 & 0.524 & 0.51 & 0.89 & 3.6 & 53.6 & 1.060 & 0.564 & 0.52 & 0.89 & 4.0 \\
NBF(2048) & 23.2 & 0.949 & 0.516 & 0.51 & 0.90 & 6.1 & 52.9 & 1.041 & 0.555 & 0.53 & 0.89 & 6.6 \\
NBF(4096) & 23.1 & 0.943 & 0.513 & 0.52 & 0.90 & 11.0 & 52.6 & 1.029 & 0.549 & 0.53 & 0.89 & 11.8 \\
EnKF(50) & 32.1 & 1.503 & 0.782 & 0.50 & 0.89 & 0.01 & 64.0 & 1.524 & 0.776 & 0.50 & 0.89 & 0.01 \\
EnKF(100) & 31.7 & 1.484 & 0.770 & 0.51 & 0.91 & 0.01 & 63.1 & 1.506 & 0.764 & 0.51 & 0.91 & 0.01 \\
PF(8192) & -- & 4.275 & 3.285 & 0.04 & 0.10 & 0.17 & -- & 4.732 & 3.708 & 0.02 & 0.05 & 0.94 \\
PF(32768) & -- & 4.004 & 3.030 & 0.06 & 0.14 & 2.3 & -- & 4.668 & 3.639 & 0.02 & 0.05 & 4.6 \\
PF(131072) & -- & 3.611 & 2.685 & 0.09 & 0.20 & 9.2 & -- & 4.635 & 3.597 & 0.02 & 0.06 & 20.6 \\
PF(262144) & -- & 3.268 & 2.391 & 0.12 & 0.26 & 20.9 & -- & 4.579 & 3.545 & 0.03 & 0.06 & 41.3 \\
PF-OP(8192) & -- & 3.091 & 2.191 & 0.14 & 0.31 & 0.15 & -- & 4.287 & 3.195 & 0.05 & 0.12 & 0.95 \\
PF-OP(32768) & -- & 2.161 & 1.443 & 0.25 & 0.51 & 2.4 & -- & 4.192 & 3.093 & 0.05 & 0.13 & 4.9 \\
PF-OP(131072) & -- & 1.449 & 0.902 & 0.36 & 0.70 & 9.6 & -- & 4.054 & 2.964 & 0.06 & 0.15 & 21.9 \\
PF-OP(262144) & -- & 1.062 & 0.617 & 0.43 & 0.80 & 21.6 & -- & 3.952 & 2.865 & 0.07 & 0.16 & 43.8 \\
PF-Reg(8192) & -- & 2.408 & 1.701 & 0.24 & 0.47 & 0.17 & -- & 4.635 & 3.582 & 0.03 & 0.07 & 0.97 \\
PF-Reg(32768) & -- & 1.216 & 0.747 & 0.41 & 0.77 & 2.4 & -- & 4.497 & 3.450 & 0.03 & 0.08 & 4.7 \\
PF-Reg(131072) & -- & 0.877 & 0.489 & 0.48 & 0.87 & 9.4 & -- & 4.332 & 3.286 & 0.04 & 0.10 & 20.9 \\
PF-Reg(262144) & -- & 0.832 & 0.455 & 0.49 & 0.89 & 21.3 & -- & 4.158 & 3.127 & 0.05 & 0.12 & 41.9 \\
PF-RM(8192) & -- & 3.201 & 2.285 & 0.13 & 0.29 & 0.17 & -- & 4.348 & 3.255 & 0.05 & 0.11 & 0.96 \\
PF-RM(32768) & -- & 2.361 & 1.610 & 0.23 & 0.47 & 2.3 & -- & 4.227 & 3.138 & 0.05 & 0.13 & 4.6 \\
PF-RM(131072) & -- & 1.526 & 0.959 & 0.34 & 0.67 & 9.3 & -- & 4.108 & 3.017 & 0.06 & 0.14 & 20.8 \\
PF-RM(262144) & -- & 1.160 & 0.687 & 0.40 & 0.77 & 21.2 & -- & 4.051 & 2.967 & 0.06 & 0.14 & 42.0 \\
\bottomrule
\end{tabular}

\end{table}

\begin{table}[ht]
    \centering
    \scriptsize
    \setlength{\tabcolsep}{3pt}
    \caption{Pursuit--evasion in distribution, every method, averaged with equal weight over the four training-mixture policies. SEMs are at most 1.0 nats on NBF's NLL and 0.03 on CRPS.}
    \label{tab:pe-ladder}
    \begin{tabular}{lrrrrrrrrrrrr}
\toprule
 & \multicolumn{4}{c}{$k = 1$} & \multicolumn{4}{c}{$k = 3$} & \multicolumn{4}{c}{$k = 5$} \\
\cmidrule(lr){2-5} \cmidrule(lr){6-9} \cmidrule(lr){10-13}
Method & NLL & CRPS & cov50 & cov90 & NLL & CRPS & cov50 & cov90 & NLL & CRPS & cov50 & cov90 \\
\midrule
NBF(512) & -0.01 & 1.285 & 0.54 & 0.92 & 11.6 & 1.087 & 0.59 & 0.97 & 22.0 & 0.937 & 0.62 & 0.96 \\
NBF(1024) & -0.05 & 1.284 & 0.54 & 0.92 & 10.7 & 1.077 & 0.58 & 0.97 & 20.8 & 0.929 & 0.62 & 0.97 \\
NBF(2048) & -0.07 & 1.280 & 0.54 & 0.93 & 10.5 & 1.070 & 0.57 & 0.97 & 20.3 & 0.926 & 0.61 & 0.97 \\
PF(8192) & -- & 1.150 & 0.49 & 0.88 & -- & 1.889 & 0.15 & 0.31 & -- & 1.754 & 0.06 & 0.14 \\
PF(32768) & -- & 1.090 & 0.50 & 0.90 & -- & 1.698 & 0.23 & 0.46 & -- & 1.685 & 0.09 & 0.20 \\
PF-Reg(8192) & -- & 1.234 & 0.41 & 0.76 & -- & 1.928 & 0.15 & 0.32 & -- & 1.774 & 0.07 & 0.16 \\
PF-Reg(32768) & -- & 1.157 & 0.43 & 0.77 & -- & 1.678 & 0.24 & 0.47 & -- & 1.698 & 0.10 & 0.22 \\
EnKF(50) & 22.4 & 1.503 & 0.47 & 0.80 & 91.2 & 1.114 & 0.46 & 0.80 & 1346.4 & 1.026 & 0.41 & 0.75 \\
EnKF(100) & 5.25 & 1.500 & 0.50 & 0.82 & 15.0 & 1.025 & 0.54 & 0.83 & 31.1 & 0.897 & 0.50 & 0.81 \\
GMF(16) & 35.0 & 1.692 & 0.46 & 0.96 & 2918.7 & 1.380 & 0.39 & 0.72 & 8644.9 & 1.229 & 0.37 & 0.70 \\
GMF(64) & 26.8 & 1.697 & 0.45 & 0.94 & 958.6 & 1.320 & 0.41 & 0.75 & 3703.4 & 1.199 & 0.38 & 0.72 \\
\bottomrule
\end{tabular}

\end{table}

\begin{table}[ht]
    \centering
    \scriptsize
    \setlength{\tabcolsep}{3pt}
    \caption{Pursuit--evasion in distribution on each training-mixture policy. The NLL SEM, at most 0.4 nats for NBF and 0.6 for the EnKF.}
    \label{tab:pe-policies}
    \begin{tabular}{lrrrrrrrr}
\toprule
 & \multicolumn{2}{c}{goal} & \multicolumn{2}{c}{random walk} & \multicolumn{2}{c}{patrol} & \multicolumn{2}{c}{evasive} \\
\cmidrule(lr){2-3} \cmidrule(lr){4-5} \cmidrule(lr){6-7} \cmidrule(lr){8-9}
Method & NLL & CRPS & NLL & CRPS & NLL & CRPS & NLL & CRPS \\
\midrule
\multicolumn{9}{l}{$k = 1$} \\
NBF(2048) & -0.56 & 0.489 $\pm$ 0.005 & 1.14 & 1.585 $\pm$ 0.002 & 0.18 & 0.781 $\pm$ 0.017 & -1.04 & 2.265 $\pm$ 0.005 \\
PF(32768) & -- & 0.305 $\pm$ 0.017 & -- & 1.583 $\pm$ 0.027 & -- & 0.253 $\pm$ 0.018 & -- & 2.220 $\pm$ 0.026 \\
EnKF(100) & 3.17 & 0.750 $\pm$ 0.029 & 5.14 & 1.768 $\pm$ 0.036 & 3.64 & 1.001 $\pm$ 0.025 & 9.04 & 2.481 $\pm$ 0.038 \\
\midrule
\multicolumn{9}{l}{$k = 3$} \\
NBF(2048) & 10.4 & 0.852 $\pm$ 0.012 & 12.5 & 1.128 $\pm$ 0.002 & 11.1 & 0.898 $\pm$ 0.008 & 7.88 & 1.403 $\pm$ 0.003 \\
PF(32768) & -- & 0.983 $\pm$ 0.037 & -- & 1.945 $\pm$ 0.038 & -- & 1.197 $\pm$ 0.040 & -- & 2.666 $\pm$ 0.061 \\
EnKF(100) & 13.5 & 0.639 $\pm$ 0.013 & 15.7 & 1.179 $\pm$ 0.016 & 12.9 & 0.747 $\pm$ 0.014 & 18.2 & 1.536 $\pm$ 0.021 \\
\midrule
\multicolumn{9}{l}{$k = 5$} \\
NBF(2048) & 19.9 & 0.852 $\pm$ 0.007 & 23.4 & 0.952 $\pm$ 0.001 & 21.0 & 0.788 $\pm$ 0.003 & 17.0 & 1.114 $\pm$ 0.002 \\
PF(32768) & -- & 1.174 $\pm$ 0.027 & -- & 1.879 $\pm$ 0.027 & -- & 1.291 $\pm$ 0.029 & -- & 2.397 $\pm$ 0.042 \\
EnKF(100) & 24.9 & 0.567 $\pm$ 0.009 & 29.2 & 1.024 $\pm$ 0.014 & 23.3 & 0.648 $\pm$ 0.011 & 47.0 & 1.347 $\pm$ 0.017 \\
\bottomrule
\end{tabular}

\end{table}

\section{Code}
\label{app:code}

All source code to reproduce these experiments will be made publicly available on \texttt{GitHub} upon publication.

\end{document}